\pdfoutput=1
\documentclass[11pt]{article}

\usepackage{acl2024}

\usepackage{times}
\usepackage{latexsym}

\usepackage[T1]{fontenc}
\usepackage[utf8]{inputenc}

\usepackage{microtype}

\usepackage{inconsolata}

\usepackage[utf8]{inputenc}
\usepackage{booktabs,multirow}
\usepackage{enumitem}
\usepackage{kotex} % Menu - Tex Live version - 2021 로 설정필요
\usepackage{color}
\usepackage{rotating}
\usepackage{amsmath}
\usepackage{amsfonts}
\usepackage{pifont}
\usepackage{array}
\usepackage{amssymb}
\usepackage[linesnumbered,algoruled,boxed,lined]{algorithm2e}

\newcommand{\ie}{{\it i.e.}}
\newcommand{\eg}{{\it e.g.}}

\newcommand{\es}{\textcolor{orange}}

\newcommand{\argmax}{\operatornamewithlimits{argmax}}

\usepackage{soul}
\usepackage{xcolor}
\definecolor{customorange}{HTML}{C55A11}

\usepackage{subcaption}
\usepackage{multirow}
\title{Multi-Granularity Guided Fusion-in-Decoder}
\author{Eunseong Choi, Hyeri Lee, Jongwuk Lee\thanks{\ \ Corresponding author} \\
        Sungkyunkwan University, Republic of Korea\\  
        \texttt{\{eunseong, bluepig94, jongwuklee\}@skku.edu}}

\begin{document}
\maketitle
\begin{abstract}
In Open-domain Question Answering (ODQA), it is essential to discern relevant contexts as evidence and avoid spurious ones among retrieved results. The model architecture that uses concatenated multiple contexts in the decoding phase, \ie, Fusion-in-Decoder, demonstrates promising performance but generates incorrect outputs from seemingly plausible contexts. To address this problem, we propose the \emph{\textbf{M}ulti-\textbf{G}ranularity guided \textbf{F}usion-\textbf{i}n-\textbf{D}ecoder (\textbf{MGFiD})}, discerning evidence across multiple levels of granularity. Based on multi-task learning, MGFiD harmonizes passage re-ranking with sentence classification. It aggregates evident sentences into an \emph{anchor vector} that instructs the decoder. Additionally, it improves decoding efficiency by reusing the results of passage re-ranking for \emph{passage pruning}. Through our experiments, MGFiD outperforms existing models on the Natural Questions (NQ) and TriviaQA (TQA) datasets, highlighting the benefits of its multi-granularity solution.

% Retrieval-augmented generation (RAG) is increasingly recognized as a fundamental approach to empowering large language models. In Open-domain Question Answering (ODQA), a prominent task where RAG shows strong potential, the use of embeddings from multiple documents at once in the decoding phase has become a leading methodology. However, there remains room for improvement in correctly reading and interpreting retrieval results. This paper introduces a novel method, \emph{\textbf{M}ulti-\textbf{G}ranularity Guided \textbf{F}usion-\textbf{i}n-\textbf{D}ecoder (MGFiD)}, which incorporates a multi-granularity-based evidence discernment identifying sentence- and passage- level evidence to grant discernment to the model. Instead of existing heuristic labels for evidence, it utilizes LLMs to classify evidence. Because the model can assess evidentiality, it can increase efficiency by only passing documents that meet a certain threshold. Experiment results show that MGFiD outperforms existing models on two widely used datasets in the ODQA task, and the ability to discern multi-granular evidence can help generate more accurate answers.
\end{abstract}

\section{Introduction}

\begin{figure}[t]
\includegraphics[width=1.0\linewidth]{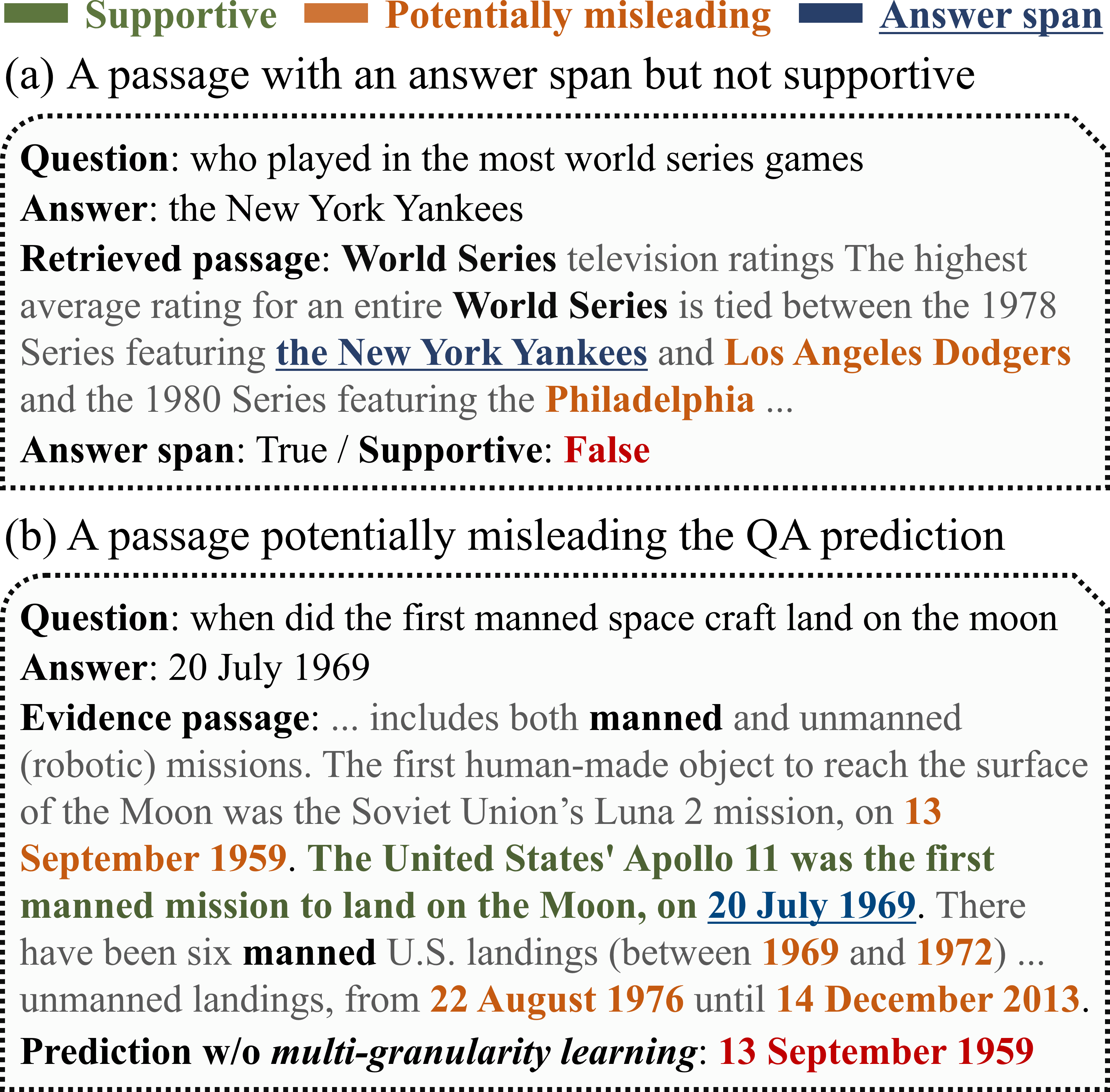}
\caption{Examples that may harm the QA systems. Black \textbf{Bold} terms in the passages are overlapped with the question.
(a) The passage is not supportive while containing a correct answer span.
(b) Confusing sentences within the passage mislead model prediction.
}\label{fig:NotEvident}
\vskip -0.1in
\end{figure}

Open-domain question answering (ODQA)~\cite{acl/ChenFWB17/DrQA} is a challenging task that requires deriving factual responses from a vast knowledge corpus without relying on explicit evidence, \ie, the evidence context is not given. Recently, retrieval-augmented generation (RAG)~\cite{nips/LewisPPPKGKLYR020/RAG} has emerged to combine the retrieval of relevant information with response generation.

Exemplified by the \emph{retriever-reader} architecture~\cite{acl/ChenFWB17/DrQA, acl/LeeCT19/ORQA, icml/GuuLTPC20/REALM}, RAG effectively addresses ODQA. The retriever first pinpoints the most relevant passages using the question as a query. Subsequently, the reader extracts or generates a response using the question and the relevant passages. It generally allows us to perform a decoupled optimization for the retriever or the reader. In this paper, we mainly focus on optimizing the reader.

To improve the reader, existing studies~\cite{eacl/IzacardG21/FiD, naacl/Asai0H22/EvidentialityQA, acl/WangY023/RFiD} focus on addressing two questions: (i) how to effectively use the evidence in multiple passages, and (ii) how to improve the discrimination in dealing with spurious passages.

\textbf{Multi-passage reader.} As the representative model, Fusion-in-Decoder (FiD)~\cite{eacl/IzacardG21/FiD} using a generative text-to-text model~\cite{JMLR/Raffel2020/T5} is an effective multi-passage reader to aggregate evidence across multiple passages. It first encodes multiple pairs of a question and a relevant passage at the encoder. Then, it generates an answer using a cross-attention mechanism over concatenated embeddings at the decoder. One limitation of the FiD architecture is inefficiency due to the intensive cross-attention operations performed on the concatenated matrix. To mitigate this, some studies have proposed either shortening the input length~\cite{sigir/HofstatterC0Z23/FiD-Light, acl/Yu0F0WXRY022/KG-FiD} or omitting some layers in the decoder~\cite{acl/JongZAFSSC23/FiDO}. More importantly, the standard FiD model often struggles with handling spurious passages that degrade the accuracy of generating the answer~\cite{naacl/Asai0H22/EvidentialityQA}.

\textbf{Multi-task reader.} Several studies~\cite{acl/Yu0F0WXRY022/KG-FiD, emnlp/Ju00Z022/GRAPE, emnlp/LakhotiaPGYMI21/FiD-Ex, sigir/HofstatterC0Z23/FiD-Light, acl/WangY023/RFiD} have attempted to address handling spurious passages by employing multi-task learning. It aims to improve the reader's discernment regarding the evidentiality of the retrieved passages, thereby achieving robustness against spurious ones. \citet{acl/Yu0F0WXRY022/KG-FiD} and \citet{emnlp/Ju00Z022/GRAPE} proposed to incorporate information from factual triplets contained in the knowledge graph. Another solution is to employ passage labels to discern spurious passages in the FiD architecture. \citet{emnlp/LakhotiaPGYMI21/FiD-Ex, sigir/HofstatterC0Z23/FiD-Light, acl/WangY023/RFiD} determined the rationality of passages based on whether they contain an answer span. Although learning signals from answer span inclusions has been proven effective, it may lead to false positive passages, producing sub-optimal results. Furthermore, existing multi-task readers face challenges in identifying the key sentences within the passage.

%While existing multi-task readers effectively have improved their abilities to distinguish supportive passages, they still lack identifying key sentences within the passage.

We argue that relying solely on answer spans or identifying evident passages is insufficient to determine the evidence. Figure~\ref{fig:NotEvident} illustrates two plausible scenarios, highlighting the limitations of existing methods using either the answer spans or passage-level evidentiality. In Figure~\ref{fig:NotEvident}(a), the mere presence of the answer span in the passage does not guarantee relevance for the question. More importantly, Figure~\ref{fig:NotEvident}(b) shows that a model trained primarily on aggregating evidence across passages generates an incorrect answer, and there is a need to distinguish complex and confusing sentences.

%More importantly, Figure~\ref{fig:NotEvident}(b) shows that a model trained primarily on aggregating evidence across passages needs to distinguish complex and confusing sentences containing incorrect answers.

This paper aims to discern evidence in coarse- and fine-grained textual information, \ie, passages and sentences, and utilize the byproduct from multi-task learning to enhance the model's performance. To this end, we propose a novel model called \emph{\textbf{M}ulti-\textbf{G}ranularity guided \textbf{F}usion-\textbf{i}n-\textbf{D}ecoder (\textbf{MGFiD})}. Specifically, the key idea behind MGFiD is two-fold. (i) We train the FiD to distinguish evidentiality using multi-task learning to minimize the influence of false contexts during answer generation. In this process, we employ both passage- and sentence-level contexts to account for evidentiality in multi-granularity contexts. Since it is expensive to label gold passages, we use the ranking abilities of language models~\cite{DBLP:conf/emnlp/RankGPT} to filter out irrelevant contexts for the question. (ii) We reuse auxiliary information from multi-task learning to improve accuracy and efficiency. We generate an \emph{anchor vector} derived from sentence-level classification and infuse it into the [BOS] token used in the decoder. Since the anchor vector indicates a significant feature for relevant sentences, it helps the decoder generate the correct answer. Furthermore, we employ passage-level re-ranking results to prune less supportive passages, improving the efficiency in the decoding phase.

To summarize, the key contributions of this paper are as follows. (i) We introduce the evidentiality of the FiD using multi-granularity contexts. (ii) We utilize LLMs to generate pseudo-labels for supportive passages in ODQA task. (iii) We reuse multi-granularity contexts to improve accuracy and efficiency further using an anchor vector in the decoder and thresholding-based passage pruning. (iv) Through our experiments on two benchmark datasets, we show that MGFiD improves the original FiD by more than 3.5\% and 1.0\% in Exact Match on Natural Questions, and TriviaQA, outperforming the other baselines.

\section{Related Work} \label{sec:related_work}
\noindent

We briefly review existing studies for improving the FiD~\cite{eacl/IzacardG21/FiD} architecture in two key aspects: \emph{accuracy} and \emph{efficiency}.

\subsection{Encoding Evident Passages}\label{sec:related_evidence}
Several works~\cite{emnlp/LakhotiaPGYMI21/FiD-Ex, sigir/HofstatterC0Z23/FiD-Light, acl/WangY023/RFiD} introduce multi-task learning to endow the model with discriminative ability, \ie, the capacity to identify spurious passages. \citet{emnlp/Ju00Z022/GRAPE} incorporates informative contexts in the knowledge graph with the reader. It extracts entity embeddings from the intermediate layer and combines them with graph knowledge fused through GNN. While relational information from the knowledge graph is helpful, it requires external sources. Another direction is to use heuristic rationale in multi-task learning. \citet{emnlp/LakhotiaPGYMI21/FiD-Ex} proposed a special sentence marker token to enable the decoder to generate a marker corresponding to the grounds along with the answer. \citet{acl/WangY023/RFiD} introduced a binary classifier to determine whether each passage is supportive between the encoder and decoder. Defining rational passages is based on the answer span. As it does not guarantee the evidentiality of the passage, ~\citet{naacl/Asai0H22/EvidentialityQA} pointed out this limitation and suggested a classifier for mining pseudo-evidentiality labels. However, it still requires expensive annotations to train the classifier, and labeling with a partially trained model can be affected by the model's memorization.

\subsection{Decoding Efficiency}\label{sec:related_efficiency}

The decoding step, mainly due to the large key-value matrix, is the most time-consuming phase in the FiD architecture during inference. Simply reducing the number of FiD inputs is not as optimal as reducing the decoder input alone~\cite{sigir/HofstatterC0Z23/FiD-Light}. Previous work has reduced the burden on the decoder by giving only necessary information. ~\citet{sigir/HofstatterC0Z23/FiD-Light} reduce the length of each encoded query-passage pair to the first few vectors. Compressing the amount of information fed to the decoder can significantly improve inference efficiency while slightly reducing effectiveness. ~\citet{acl/JongZAFSSC23/FiDO} removes most cross-attention layers and employs multi-query attention to reduce the cost of the decoder. \citet{acl/Yu0F0WXRY022/KG-FiD} takes intermediate layer representation for passage re-ranking and improves efficiency by passing only the high-ranked passages to the decoder. However, using a fixed number of passages is problematic as it assumes that the number of supporting documents is constant, whereas they vary.
\begin{figure*}[t]
\includegraphics[width=1.0\linewidth]{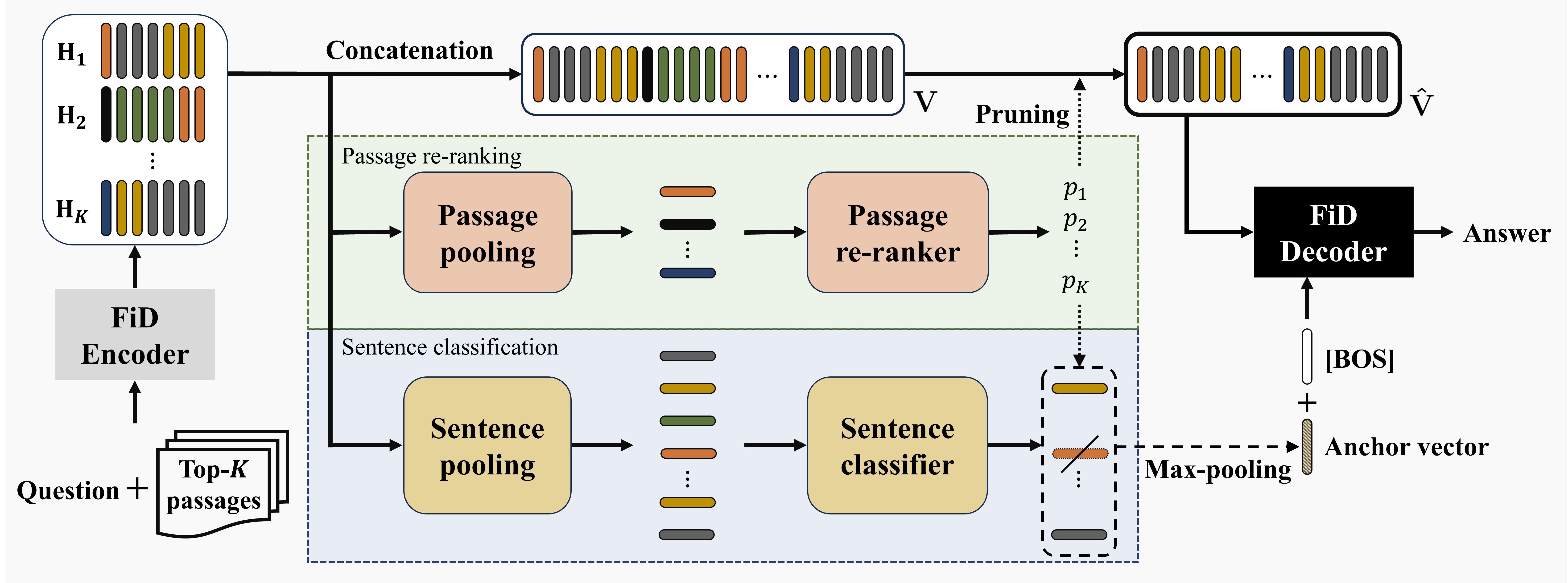}
\caption{The MGFiD framework incorporates multi-task learning for answer generation, leveraging passage re-ranking to identify coarse-grained evidence and sentence classification for fine-grained evidence. It utilizes the outcomes of these tasks—threshold-based masking from passage re-ranking and anchor embedding from sentence classification—to enhance both efficiency and effectiveness in the answer generation process.}\label{fig:architecture}
\vskip -0.1in
\end{figure*}

\section{Proposed Method}

In this section, we first outline our method for multi-task learning, which integrates generating answers and determining their evidence at different levels of granularity, \ie, passages and sentences (Section~\ref{sec:multi_task}). Second, leveraging sentence-level predictions, we introduce an anchor vector to provide a rationale signal to the decoder (Section~\ref{sec:anchor_sent}). We then present threshold-based pruning using passage-level scores to enable efficient decoding (Section~\ref{sec:masking}). Lastly, we describe the process of generating pseudo-labels for supportive passages (section~\ref{sec:ev_labeling}).

\subsection{Learning Multi-granularity Contexts}\label{sec:multi_task}

%, which is known to effectively aggregate the evidence across contexts

\textbf{Answer generation.} We adopt the standard FiD~\cite{eacl/IzacardG21/FiD} architecture as our base model. The FiD encoder takes as input the top-$K$ retrieved passages $\text{P}_{\text{q}}=[\text{p}_1, \text{p}_2, ..., \text{p}_K]$ for the question $\text{q}$. Each $\text{p}_i$ is prepended with $\text{q}$, and the FiD encoder outputs the token embeddings $\mathbf{H_i}$, which are then concatenated to obtain $\mathbf{V}$.
\begin{equation}
\begin{gathered}
\mathbf{H}_i=\text{FiD}_{\text{encoder}}\left(\text{q}+\text{p}_i\right) \in\mathbb{R}^{L \times d}, \\
\mathbf{V}=\left[\mathbf{H}_1 ; \mathbf{H}_2 ; \ldots ; \mathbf{H}_K\right] \in\mathbb{R}^{(K \times L) \times d}.
\end{gathered}
\end{equation}

Here, $L$ denotes the maximum sequence length, and $d$ denotes the hidden dimension. The FiD decoder utilizes $\textbf{V}$ as the key-value matrix to generate the answer auto-regressively. When $T$ is the target sequence, the loss function is as follows:
\begin{equation}
\mathcal{L}_{\text{gen}} = -\sum_{t=1}^{T} \log p(\hat{y}_t \mid y_{<t}, \mathbf{V}).
\end{equation}

%of the answer generation is defined

\vspace{1.0mm}
\noindent
\textbf{Passage re-ranking}. When the original FiD is solely trained on answer generation, it tends to predict incorrect answers from plausible passages, \eg, passages with word overlap to the question but not supportive. To mitigate this, we account for re-ranking the evidentiality of passages. Inspired by \cite{corr/abs-1901-04085/monoBERT}, we obtain the evidence embedding $\mathbf{e}_i\in \mathbb{R}^{1 \times d}$ by passing the first token embedding of each pair to the projection layer. Specifically, let $\textbf{h}^{j}_{i} \in \mathbb{R}^{1 \times d}$ be $j$-th token embedding of the $\textbf{H}_i$, which denotes token embeddings of the question and $i$-th passage; we pass $\textbf{h}^{0}_{i} \in \mathbb{R}^{1 \times d}$ through the projection layer $\mathbf{W}_p \in \mathbb{R}^{d \times d}$. Then, a single-layer neural network $\mathbf{W}_r \in \mathbb{R}^{1 \times d}$ takes $\mathbf{e}_i$ to predict the logit for each passage. A softmax function is applied to $K$ logits to get a probability $p_i$ for the question and $i$-th passage pair.
\begin{equation}\label{eq_evidence}
p_i = \text{softmax}(\mathbf{e}_i \mathbf{W}^{\top}_r), \text{where} \ \mathbf{e}_i = \mathbf{h}^0_i \mathbf{W}_p.
\end{equation}

Using the probability $p_i$, the loss function with negative log likelihood for passage re-ranking $\mathcal{L}_{\text{passage}}$ is defined by:

\begin{equation}\label{eq_l_evidence}
\mathcal{L}_{\text{passage}} = -\frac{1}{|\mathcal{P}|} \sum_{pos \in \mathcal{P}} \log (p_{pos}).
\end{equation}

$\mathcal{P}$ denotes a set of indices for positive passages corresponding to the question. Here, $\mathcal{L}_{\text{passage}}$ highlights passages containing evidence and guides the decoder to focus on considering more relevant passages in generating the answer.

%The loss function is defined by:

We adopt a listwise loss function rather than a pointwise because it makes sense to focus on relative evidentiality between $K$ passages. Furthermore, $p_i$, which represents the relative importance of each passage, is subsequently used for threshold-based masking for efficient decoding (Section~\ref{sec:masking}).

% A critical part of this methodology is the quality of the labels. However, gold context labels are often provided in a limited way in ODQA. While prior work~\cite{acl/WangY023/RFiD} has shown improvement using the signal from the answer span, we propose to leverage the ranking capabilities of LLMs~\cite{corr/abs-2304-09542/RankGPT} to label the evidentiality of passages. Specifically, we use a triplet of a question, a passage, and candidate answers to determine evidentiality. This process is costly for every triplet, but as we want to filter out spurious passages, we reduce the cost by focusing only on those that contain the answer span. To assess the effectiveness of LLMs in the labeling task, we validate indirectly by observing the filtering rate based on DPR rank. (Refer to the details in Appendix~\ref{sec:exp_eval_on_filtering}.)

\vspace{1.0mm}
\noindent
\textbf{Sentence classification.} To leverage evidentiality in nuanced text, we deal with a fine-grained sentence-level task. Previous work~\cite{cikm/0001HGCHZ23/GranCATs} has combined different granularities to enrich the global semantics, suggesting that the information that can be captured at different levels of granularity is different. This implies that the coarse-grained semantics alone is insufficient to determine which sentences are support sentences. Therefore, multi-granularity evidentiality helps improve discrimination.

We enhance the model by learning local evidence from fine-grained sentences. Specifically, the sentence classifier takes a sentence embedding as input to predict whether the answer span is in the sentence or not. Since we need to distinguish between sentences, not their relative importance, it is designed as a simple classification task rather than a ranking task. The $n$-th sentence embedding of the $i$-th passage $\mathbf{s}^{n}_{i} \in \mathbb{R}^{1 \times d}$ is expressed as the average of token embeddings projected by $\textbf{W}_p \in \mathbb{R}^{d \times d}$. The loss function is defined after the sentence classification layer $\mathbf{W}_s\in \mathbb{R}^{d\times2}$.
\begin{equation}\label{eq_sentence}
\begin{gathered}
\mathbf{s}^{n}_{i} = \text{mean-pooling} \left( \left\{ \mathbf{h}_i^j \mathbf{W}_p \right\}_{j=a_n}^{b_n} \right), \\
\mathcal{L}_{\text{sentence}}^{n,i} = \text{Focal}\left(y_{i}^{n}, \mathbf{s}_{i}^{n} \mathbf{W}_s\right). \\
\end{gathered}
\end{equation}

%\mathbf{s}^{n}_{i} = \text{mean-pooling} \left( \{ \mathbf{W}_p \mathbf{h}_i^j \}_{j=a_n}^{b_n} \right), \\
% \mathbf{s}^{n}_{i} = \text{mean-pooling} \left( \bigoplus_{j=a_n}^{b_n} \mathbf{W}_p \mathbf{h}_i^j \right), \\
% \mathbf{s}^{n}_{i} = \text{mean-pooling}_{j=a_n}^{b_n} \mathbf{W}_p \mathbf{h}_i^j, \\\textbf{}
% \begin{equation}
% \begin{gathered}
% sent_{cls}\ equation, \textbf{h}^{j}_{i} \\
% \mathcal{L}_{sent}\ equation,
% \end{gathered}
% \end{equation}

Let $a_{n}$ and $b_{n}$ be the start and end token indices for the $n$-th sentence. Focal$(\cdot, \cdot)$ is the focal loss function~\cite{iccv/LinGGHD17/focal} that addresses class imbalance, with $y_{i}^{n}$ as the label indicating the presence of the answer span in the $n$-th sentence of the $i$-th passage. $\mathcal{L}_{\text{sentence}}$ is calculated as the average of all sentences in the batch.

Since the passage has been validated by LLM, using the answer span information within the passage gets more accurate. We label all sentences as negative if the passage was deemed unsupportive, regardless of the answer span. Finally, the multitask learning loss to train MG-FiD is computed as a linear combination of the three loss functions.
\begin{equation}\label{eq_loss_all}
\mathcal{L} = \mathcal{L}_{\text{gen}} + \lambda_{1} \cdot \mathcal{L}_{\text{passage}} + \lambda_{2} \cdot \mathcal{L}_{\text{sentence}},
\end{equation}
\noindent
where $\lambda_{1}$ and $\lambda_{2}$ are hyperparameters that adjust the influence of passage re-ranking and sentence classification, respectively.

% As illustrated in Figure~\ref{fig:sentonly}, the sentence with the answer span may not provide enough contextual evidence to inform the model, leading to difficulties in identifying supportive passages.
Figure~\ref{fig:sentonly} illustrates the result when only sentence classification multi-task learning is performed. It fails to learn the relationship between the question and the coarse-grained passage, and may generate the answer from the plausible passages.
That is, focusing solely on sentences can limit broader semantic understanding. In detail, since the passage is prepended with the question, the first token embedding $\mathbf{h}_{i}^{0}$ used for the passage embedding is always the \emph{"question"} token embedding. On the other hand, a sentence uses the start and end token indices of each sentence in the passage, so there is no overlap between the passage and sentence embeddings.

% We underline the importance of our multi-granularity approach: i) 
% the sentence classification captures fine-grained evidence,
% potentially impacts the model's ability to generalize beyond isolated facts.
% as illustrated in Figure~\ref{fig:sentonly}. This underlines the importance of our multi-granularity approach.

\begin{figure}[t]
\includegraphics[width=1.0\linewidth]{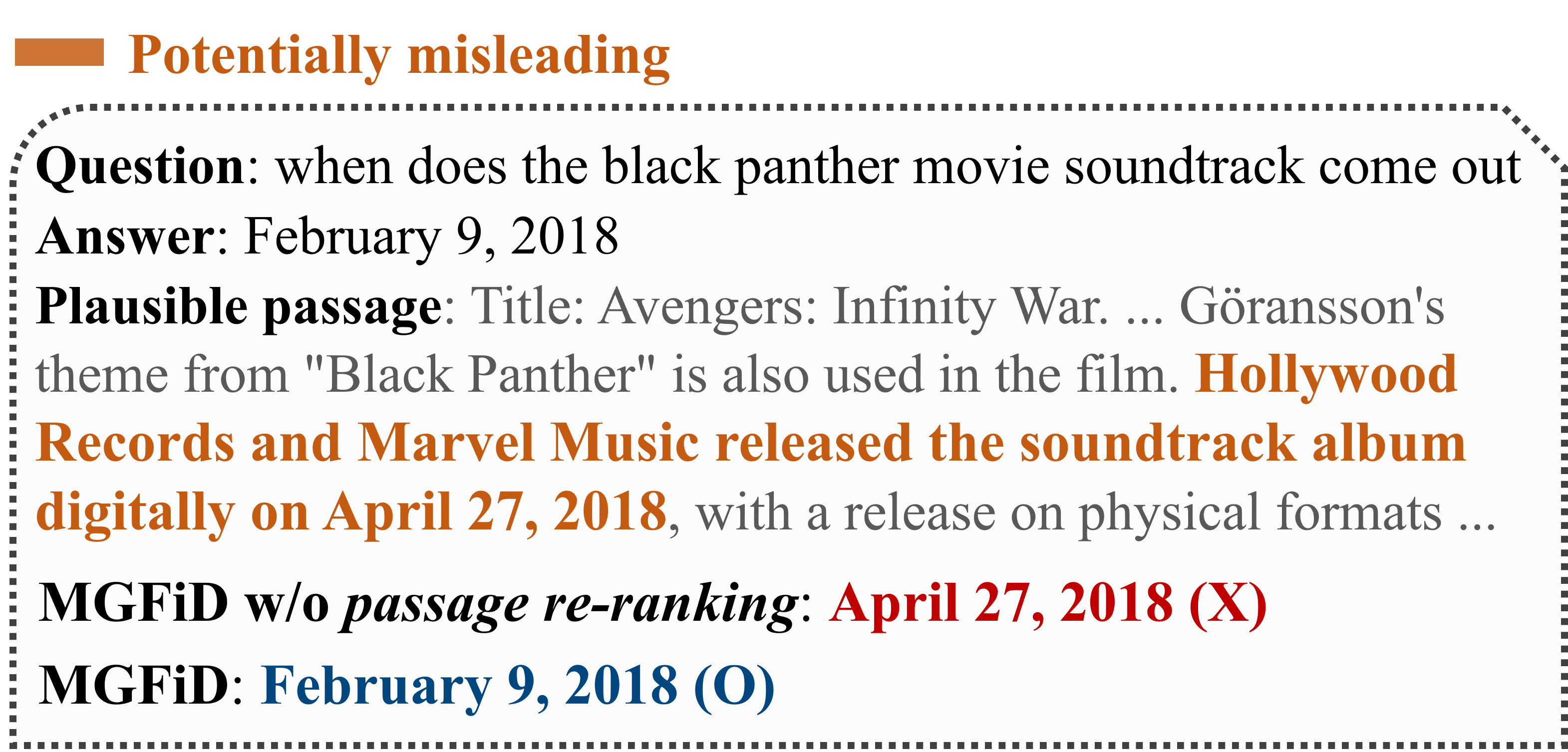}
\caption{Learning solely from sentences may lead to a lack of understanding of the broader context.}\label{fig:sentonly}
\vskip -0.1in
\end{figure}

\iffalse
\begin{figure}[t]
\includegraphics[width=1.0\linewidth]{Figures/figure_sentonlynotenough_v2.png}
\caption{Example of insufficient contextual evidence in the sentence. While having proper clues for the prediction, learning solely from sentences may lead to a lack of understanding of the broader context.}\label{fig:sentonly}
\vskip -0.1in
\end{figure}
\fi
% This can cause a model to not focus on the clues presented in these sentences.

\subsection{Incorporating an Anchor Vector}\label{sec:anchor_sent}

While our model is trained to discern at multiple levels of granularity, thereby highlighting evidential passages and sentences, how the decoder leverages this highlighted information remains unexplored. To deal with it, we align the multi-task of identifying supportive contexts with the answer generation. Specifically, we initiate the decoder with an \emph{anchor vector}, aiming for a more focused and effective processing of relevant contexts. We leverage a set of sentences positively predicted by the sentence classifier and deal with them as an extractive summarization across multiple passages. The anchor vector, denoted as  $\textbf{e}_{\text{anchor}} \in \mathbb{R}^{1 \times d}$, is obtained by performing the max-pooling operation on these positively predicted sentence embeddings, which then employed to couple the generation with the multi-task learning.

We first obtain a set of sentence embeddings $S$ that are predicted as positives by the sentence classifier across the $K$ passages as follows:
% \vskip -0.05in
\begin{equation}\label{eq_anchor}
\begin{aligned}
\mathcal{S} = \bigcup_{i=1}^{K} \bigg\{ \mathbf{s}_i^n \mid & \argmax(\mathbf{s}_i^n \mathbf{W}_s)= 1, \\
& \forall n \in \{1, \ldots, \text{N}_i\} \bigg\},
\end{aligned} \\
% \mathcal{S} = \bigcup_{i=1}^{K} \left\{ \mathbf{s}_i^n \mid \sgn \left( \mathbf{s}_i^n \mathbf{W}_s \right) = 1, \forall n \in \{1, \ldots, \text{N}_i\} \right\}, \\
\end{equation}
% \vskip -0.05in
Here, $\text{argmax}(\cdot)$ is applied to a two-dimensional vector for each of $\text{N}_i$ sentences in $\text{p}_i$, returning zero for negative and one for positive. As we collect the sentence embeddings that are predicted to be positive, we then apply max-pooling over $\mathcal{S}$ to obtain the anchor vector to capture the most salient evidence.
\begin{equation}\label{eq_anchor}
\mathbf{e}_{\text{anchor}} = \text{max-pooling} \left( \mathcal{S} \right).
\end{equation}

Lastly, we add the anchor vector to the existing [BOS] token embedding, allowing the decoder to use the evident information in the cross-attention mechanism. Our approach differs from the existing learnable guided embedding proposed by~\citet{acl/WangY023/RFiD}. That is, we directly incorporate the fine-grained supportive embedding into cross-attention by adding it to the query token, as distinct from expecting guided embeddings to be reflected within the decoder layer.

\subsection{Pruning Passages via a Threshold}\label{sec:masking}

To improve the cross-attention cost bottleneck in the decoder, we employ a threshold-based pruning method. Specifically, we reuse the probabilities for the $K$ passages computed in the passage re-ranking task, discarding passages below a threshold $\tau$. The resulting pruned key-value matrix $\mathbf{\hat{V}}$ based on the probability $p_{i}$ in Equation~\eqref{eq_evidence} is formed as follows:

\begin{equation}
\mathbf{\hat{V}} = \bigoplus_{i=1}^{n} \mathbf{H}_i \quad \text{if } p_i > \tau.
\end{equation}

Let $\hat{K}$ be the number of passages that exceed the threshold $\tau$; we obtain the pruned key-value matrix $\mathbf{\hat{V}} \in \mathbb{R}^{(\hat{K}\times L)\times d}$. Adjusting the threshold from 0.0 to 0.1, we found it efficient yet effective at $\tau = 0.05$. As a result, MGFiD dynamically uses only the necessary evidence for each question, instead of a fixed number of passages as in the previous methods~\cite{emnlp/LeeKLPMW22/YONO, acl/Yu0F0WXRY022/KG-FiD}, thereby improving efficiency more effectively.

\begin{figure}[t]
\includegraphics[width=1.0\linewidth]{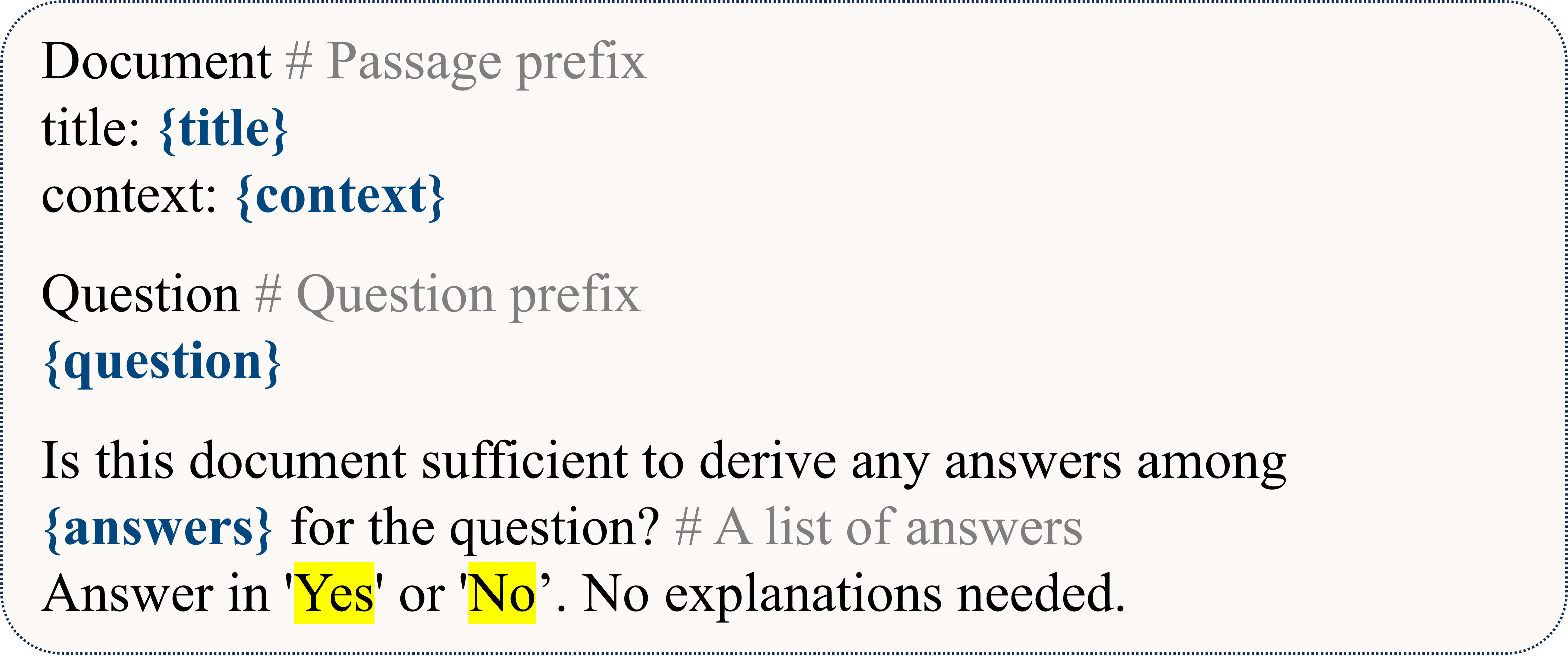}
\caption{A prompting example used for LLMs to filter out contexts that have an answer span but are not evident to the question.}\label{fig:prompt}
\vskip -0.1in
\end{figure}
\subsection{Evidence Labeling}\label{sec:ev_labeling}
A critical part of the passage re-ranking is the quality of the labels. However, gold context labels are often provided in a limited way in ODQA. While prior work~\cite{acl/WangY023/RFiD} has shown improvement using the signal from the answer span, we propose to leverage the ranking capabilities of Large Language Models (LLMs)~\cite{DBLP:conf/emnlp/RankGPT}. Specifically, we use large language models to generate pseudo-labels according to the evidentiality of passages. The prompt shown in Figure~\ref{fig:prompt} instructs the LLMs to identify a passage if it is sufficient to answer the question. While it would be a burden to generate pseudo-labels for every triplet candidate, \ie, a question, answers, and a passage, we reduce the cost by focusing only on those that contain the answer span. To assess the effectiveness of LLMs in the labeling task, we validate them indirectly by observing the filtering rate based on the retrieval~\cite{iclr/IzacardG21/FiD-KD} rank. (Refer to the details in Appendix~\ref{sec:exp_eval_on_filtering}.)

\noindent
% Figure N에서 디코더의 인풋으로 들어가는 문단이 일부 마스킹된 것을 보여준다. We simply 최소 한 개 이상의 문서가 passing 될 수 있도록 set $\tau$ to 1/$K$. We modify the attention mask for the batch process during training.
%학습과정에서 디코더는 cross-attention을 수행할 때 attention-mask가 씌워진 문단을 제외하고 나머지 문단만을 참고한다.

% 결과적으로, passage classifier의 두번째 단계인 마스킹의 결과는 다음과 같다.

% [마스킹 후 passage classifier 아웃풋 벡터 형태 - figure - ]

% 실제 추론 단계에서는 threshold를 넘기지 못하는 문서에 attention masking을 사용하는 대신에 해당 문서를 제외한 채 임베딩 concatnation을 진행한다.

% \iffalse
% Vanila FiD 모델은 검색결과 context를 모두 동등하게 디코더의 인풋으로 사용함으로써 그 중 \es{근거} context가 무엇인지 판별하는 것은 \es{암시적으로 답변을 생성하는 과정 속에 포함된다}. 기존 연구 RFiD는 디코더의 부담을 덜어내기 위하여 관련문단을 판별해내는 binary classification 레이어를 추가한다. Binary classifier는 각 Hk의 첫번째 토큰을 인풋으로 받아 rational/spurious를 뜻하는 prediction label ∈ {0,1}을 예측하도록 학습된다. 결과적으로, prediction label들은 updatable rationale embeddings으로 변환되어 encoder hidden states의 마지막에 추가된다.

% 인코더의 \textit{}hidden state에 임베딩을 추가함으로써 디코더에게 근거 문단에 대한 정보를 제공하는것은 디코더의 부담을 줄여줄 수 있지만, 오로지 근거 문단만을 입력으로 사용하는 것만큼은 아닐 것이다. 우리는 Passage reranker로 산출되는 점수를 기준으로 일정 threshold를 넘기지 못하는 문단의 임베딩에 마스킹을 함으로써 불필요한 디코더 연산을 제거하는 것을 목표로 한다.

% 앞서 설명한 passage reranker의 아웃풋 Ri는 다음과 같다.

% [리랭킹 레이어 아웃풋 벡터 형태]

% 레이어 아웃풋은 logit의 형태이며, passage score는 여기에 softmax를 취한 값이다.
% 우리는 총 20개 문서 중에서 평균보다 높은 passage score를 가지는 문단만을 사용하기 위해 threshold를 0.05로 설정하였다.
% 결과적으로, passage classifier의 두번째 단계인 마스킹의 결과는 다음과 같다.

% [Ri에 softmax를 취하는 수식]

% [마스킹 후 passage classifier 아웃풋 벡터 형태]
% \fi

\section{Experiments Setup}\label{sec:setup}

\begin{table}[t]\small
\begin{footnotesize}
\begin{tabular}{c|ccc|c|c}
\toprule
Dataset & train & dev & test & R@20 & \# pos/q \\ \midrule
NQ      & 79,168            & 8,757           & 3,610            & 0.87               & 4.5                      \\
TQA     & 78,785            & 8,837           & 11,313           & 0.86               & 8.9                      \\ \bottomrule
\end{tabular}
\end{footnotesize}
\caption{Data statistics. \# pos/q indicates the average number of passages that have the answer span per the question. R@$K$ is one if there exists a positive among the $K$ passages and zero. R@20 and \# pos/q are for the training dataset using the retriever~\cite{emnlp/KarpukhinOMLWEC20/DPR} trained by~\citet{iclr/IzacardG21/FiD-KD}.}
\label{tab:dataset}
\end{table}
\subsection{Datasets}\label{sec:datasets}
We conduct our experiments on two benchmark datasets: Natural Questions (NQ)~\cite{tacl/KwiatkowskiPRCP19/NQ} and TriviaQA (TQA)~\cite{acl/JoshiCWZ17/TQA}. NQ comprises actual Google search queries, while TQA comprises question-answer pairs sourced from trivia and quiz-league websites. Table~\ref{tab:dataset} presents the statistical details of datasets.

\subsection{Metrics}\label{sec:datasets}
We use three metrics in our experiments. Exact Match (EM) evaluates the accuracy of the QA task by examining whether normalized predictions exactly match ground-truth answers. Recall@$K$ (R@$K$) assesses passage ranking, with a scoring one if the passage containing the answer span is among the top-$K$ passages. For sentence classification, we use the area under the ROC curve (AUC)~\cite{pr/Bradley97/AUC}, to account for class imbalance, \ie, most are negative.

\begin{table*}[t] \small
\begin{tabular}{cccccccc}
\toprule
\multicolumn{1}{c|}{\multirow{2}{*}{Model}} & \multirow{2}{*}{\begin{tabular}[c]{@{}c@{}}Multi-task\\ learning\end{tabular}} & \multirow{2}{*}{Retriever} & \multicolumn{1}{c|}{\multirow{2}{*}{\begin{tabular}[c]{@{}c@{}}Avg. \# psgs\\ in Decoder\end{tabular}}} & \multicolumn{2}{c|}{NQ (EM)}                 & \multicolumn{2}{c}{TQA (EM)} \\
\multicolumn{1}{c|}{}                       &                             &                            & \multicolumn{1}{c|}{}                                                                                   & Dev        & \multicolumn{1}{c|}{Test}       & Dev             & Test            \\ \midrule
\multicolumn{1}{c|}{FiD~\citeyearpar{eacl/IzacardG21/FiD}}            & -                           & DPR                        & \multicolumn{1}{c|}{100}                                                               & 46.5       & \multicolumn{1}{c|}{48.2}       & 64.7            & 65.0              \\
\multicolumn{1}{c|}{GRAPE~\citeyearpar{emnlp/Ju00Z022/GRAPE}}           & O                           & DPR                        & \multicolumn{1}{c|}{100}                                                                                   & -          & \multicolumn{1}{c|}{48.7}       & -               & 66.2            \\
\multicolumn{1}{c|}{FiD-KD~\citeyearpar{iclr/IzacardG21/FiD-KD}}         & -                           & FiD-KD                     & \multicolumn{1}{c|}{100}                                                                                   & 49.2       & \multicolumn{1}{c|}{50.1}       & 68.7            & 69.3            \\
\multicolumn{1}{c|}{RFiD~\citeyearpar{acl/WangY023/RFiD}}            & O                           & FiD-KD                     & \multicolumn{1}{c|}{100}                                                                                   & 50.0         & \multicolumn{1}{c|}{50.7}       & 69.6            & 69.6            \\ \midrule

\multicolumn{1}{c|}{FiD~\citeyearpar{eacl/IzacardG21/FiD}}            & -                           & DPR                        & \multicolumn{1}{c|}{25}                                                                                 & 45.3       & \multicolumn{1}{c|}{-}          & 63.2            & -               \\
\multicolumn{1}{c|}{KG-FiD~\citeyearpar{acl/Yu0F0WXRY022/KG-FiD}}          & O                           & DPR + GNN                       & \multicolumn{1}{c|}{20}                                                                                   & -          & \multicolumn{1}{c|}{49.6}       & -               & 66.7            \\
\multicolumn{1}{c|}{EvidentialityQA~\citeyearpar{naacl/Asai0H22/EvidentialityQA}} & O                           & FiD-KD                     & \multicolumn{1}{c|}{20}                                                                                 & 47.8       & \multicolumn{1}{c|}{49.8}       & 67.7            & 67.8            \\ \midrule
\multicolumn{8}{c}{\textit{Our implementations}}                                                                                                                                                                                                                                                    \\ \midrule
\multicolumn{1}{c|}{FiD~\citeyearpar{eacl/IzacardG21/FiD}}         & -                           & DPR    & \multicolumn{1}{c|}{20}                                                                & 45.3 \scriptsize $\pm$ 0.31 & \multicolumn{1}{c|}{46.3 \scriptsize $\pm$ 0.10} & 61.5 \scriptsize $\pm$ 0.12    & 62.1 \scriptsize $\pm$ 0.34  \\
\multicolumn{1}{c|}{FiD-KD~\citeyearpar{iclr/IzacardG21/FiD-KD}}         & -                           & \multirow{5}{*}{FiD-KD}    &  \multicolumn{1}{c|}{100} & 49.1 & \multicolumn{1}{c|}{50.1} &  -      & -      \\
\multicolumn{1}{c|}{FiD-KD~\citeyearpar{iclr/IzacardG21/FiD-KD}}         & -                           &     &  \multicolumn{1}{c|}{20} & 47.8 \scriptsize $\pm$ 0.16 & \multicolumn{1}{c|}{48.4 \scriptsize $\pm$ 0.31} & 67.4 \scriptsize $\pm$ 0.12      & 67.6 \scriptsize $\pm$ 0.25      \\
\multicolumn{1}{c|}{EvidentialityQA~\citeyearpar{naacl/Asai0H22/EvidentialityQA}} & O                           &                            & \multicolumn{1}{c|}{20}                                                                                   & 48.0 \scriptsize $\pm$ 0.20 & \multicolumn{1}{c|}{49.0 \scriptsize $\pm$ 0.39} & n/a             & n/a             \\
\multicolumn{1}{c|}{RFiD~\citeyearpar{acl/WangY023/RFiD}}            & O                           &                            & \multicolumn{1}{c|}{100}                                                                                   & 49.2  & \multicolumn{1}{c|}{50.4} & -      & -      \\
\multicolumn{1}{c|}{RFiD~\citeyearpar{acl/WangY023/RFiD}}            & O                           &                            & \multicolumn{1}{c|}{20}                                                                                   & 48.6 \scriptsize $\pm$ 0.29 & \multicolumn{1}{c|}{49.4 \scriptsize $\pm$ 0.53} & \underline{67.8} \scriptsize $\pm$ 0.12      & \underline{68.1} \scriptsize $\pm$ 0.20      \\ \midrule
\multicolumn{1}{c|}{MGFiD}                  & O                           & \multirow{2}{*}{FiD-KD}    & \multicolumn{1}{c|}{20}                                                                                 & \textbf{49.0} \scriptsize $\pm$ 0.21 & \multicolumn{1}{c|}{\textbf{50.1} \scriptsize $\pm$ 0.33} & \textbf{68.0} \scriptsize $\pm$ 0.09      & \textbf{68.3} \scriptsize $\pm$ 0.23      \\
\multicolumn{1}{c|}{Pruned MGFiD ($\tau$=0.05)}           & O                           &                            & \multicolumn{1}{c|}{4.8 / 7.7}                                                                                & \underline{48.8} \scriptsize $\pm$ 0.20      & \multicolumn{1}{c|}{\underline{49.7} \scriptsize $\pm$ 0.52}      & \underline{67.8} \scriptsize $\pm$ 0.07            & \textbf{68.3} \scriptsize $\pm$ 0.16            \\ \bottomrule
\end{tabular}
\caption{Performance comparison between MGFiD and baseline models. Avg. \# psgs in Decoder for Pruned MGFiD is the average number of passages passed to the decoder in NQ  /  TQA, respectively. {\scriptsize $\pm$} indicates the standard deviation of 5 runs. The best result among the models using $K=20$, which is the number of retrieved passages used in the encoder, is marked \textbf{bold}, and the second best is \ul{underlined}.}
\label{tab:main}
\end{table*}

\subsection{Baselines}\label{sec:baselines}

We compare MGFiD with several baselines. FiD~\cite{eacl/IzacardG21/FiD} is the first work that utilizes concatenated passage embedding at the decoder. FiD-KD~\cite{iclr/IzacardG21/FiD-KD} improves the performance of the retriever with the aggregation capability of FiD. GRAPE~\cite{emnlp/Ju00Z022/GRAPE} exploits the relationships of triplets in a knowledge graph. RFiD~\cite{acl/WangY023/RFiD} performs multi-task learning by using the answer span and proposes learnable embedding to guide the decoder. To be concise, the difference between anchor vector and guide embedding is twofold: 1) Anchor vector expects a combination of significant evidence, unlike the predicted binary label for guide embedding. 2) Guide embedding is used in the same way as other token embeddings, while an anchor vector is explicitly added in a query token for the decoder. EvidentialityQA~\cite{naacl/Asai0H22/EvidentialityQA} adopts an additional decoder for evidence classification and proposes a classifier for evidence labeling to perform multi-task learning.

\subsection{Implementation Details}\label{sec:implementation}
As a backbone model, we initialize the model t5-base~\cite{JMLR/Raffel2020/T5}. Due to the computing cost, we mainly use the top-20, \ie, $K=20$, retrieved results provided by FiD-KD~\footnote{\url{https://github.com/facebookresearch/FiD}}. We use Adam~\cite{iclr/KingmaB14/adam} as the optimizer, with a learning rate of 1e-4. We set a batch size of 2 and an accumulation step of 16 to imitate a large batch. We set $\lambda_{1}$ for passage ranking loss to 0.5 and $\lambda_{2}$ for sentence classification to 1. The $\alpha$ for focal loss~\cite{iccv/LinGGHD17/focal}, which we omit for readability in the equation~\ref{eq_sentence}, is set to 0.95, and the $\tau$ for threshold-based pruning is set to 0.05. The total number of steps is set to 160k, and for every 8k, we perform an evaluation with the validation set and select the checkpoint with the highest validation score. The maximum input sequence length is set to 192 for NQ and 250 for TQA. We use NLTK library~\cite{lre/Wagner10/NLTK} to tokenize sentences in the passages. For evidence labeling, we use ChatGPT~\footnote{\url{https://chat.openai.com/chat}} and MythoMax~\footnote{\url{https://huggingface.co/TheBloke/MythoMax-L2-13B-GPTQ}}. While ChatGPT is a powerful large language model accessible via API, MythoMax can easily be used in the local GPUs. We fix the temperature to 0 and do not use sampling to ensure reproducibility. We use all the answer candidates in NQ; however, in the TQA dataset, which has many more answer candidates than NQ, we only collected answers in the top 20 passages for efficient prompting. The experiments in Table \ref{tab:main}, Table \ref{tab:label_comparison} and Figure \ref{fig:fig_varying_passages} represent the averages of five seeds, while the other experiments use a single, fixed seed. We use two NVIDIA A100 GPUs for training and inference.

For a fair comparison, we attempt to reproduce several baselines. We use the publicly available official implementations of each methodology and report the average of five runs with the same seed set with MGFiD. For EvidentialityQA~\footnote{\url{https://github.com/AkariAsai/evidentiality\_qa}}\label{EvidentialityQA_repo}~\citeyearpar{naacl/Asai0H22/EvidentialityQA}, we observed a technical issue with the TQA dataset in the official repository, where all evidence labels were incorrectly marked as 0. On the NQ test set, we got results that were lower than the original paper, while we got slightly better results on dev. Considering the standard deviation, we consider this to be a valid reproduction. FiD, FiD-KD~\footnotemark[1]~\citeyearpar{eacl/IzacardG21/FiD, iclr/IzacardG21/FiD-KD}, and RFiD~\footnote{\url{https://github.com/wangcunxiang/RFiD}}\label{RFiD_repo}~\citeyearpar{acl/WangY023/RFiD} originally used $K$ as 100, but for a fair comparison, we trained them using 20 after validating reproducibility.

\section{Results and Analysis}

\subsection{Main Results}\label{sec:exp_performance}

Table~\ref{tab:main} shows the effectiveness of our model with the baseline models on the NQ and TQA datasets. We report the results of our model and four replications averaged over five seeds, along with their standard deviations. All models in this experiment are initialized with T5-base~\cite{JMLR/Raffel2020/T5}. Note that MGFiD incorporates components to discriminate evidence, consisting merely of only a few MLP layers, which marginally increases the number of parameters by less than 1\% from the backbone model. Avg. \# psgs in Decoder, which is the number of passages passed to the decoder, is identical with the number of retrieved passages using in the encoder, \ie, top-$K$, except for $K=20$ for Pruned MGFiD and $K=100$ for KG-FiD~\cite{acl/Yu0F0WXRY022/KG-FiD}.

% 1. Better than FiD-KD
% 2. Better than RFiD & EvidentialityQA
% 3. pruned MGFiD
First, MGFiD significantly improves over the baseline models using the same retriever and the same number of passages. Compared to the original model, FiD-KD~\cite{iclr/IzacardG21/FiD-KD}, which only performs answer generation task, MGFiD improves the EM score on the test set by 3.5\% on the NQ and 1.0\% on the TQA, and is comparable to FiD-KD using 100 passages on the NQ dataset. This implies that MGFiD identifying evidence in the multi-granularity approach effectively guides the model into supportive passages to the question. 

Second, EvidentialityQA~\cite{naacl/Asai0H22/EvidentialityQA} and RFiD~\cite{acl/WangY023/RFiD} show improved performance compared to models without multi-task learning. This implies that determining evidentiality among passages enhances the quality of answer generation. Additionally, MGFiD further improves this process by integrating fine-grained, sentence-level evidence, demonstrating an improvement of 2.2\% and 1.4\% on the NQ test set over EvidentialityQA and RFiD, respectively. 

Third, MGFiD using passage pruning significantly reduces the number of passages used by 76\% on the NQ and 61.5\% on the TQA, lowering the number of passages to 4.8 and 7.7 passages. The key-value matrix in the decoder, as noted in FiD-Light~\cite{sigir/HofstatterC0Z23/FiD-Light}, is the most resource-intensive part of FiD. Despite passage pruning in MGFiD, there is only a decrease of less than 1\% in performance, indicating effective pruning of irrelevant passages. Furthermore, it maintains or even improves performance compared to other baseline models on both the NQ and TQA datasets.

%Such reduction in the decoder inputs modestly decreases performance only in NQ dataset by 0.8\%. Overall, it is more evident on NQ datasets than on TQA. This tendency is also seen in existing multi-tasking models, i.e., RFiD, and EvidentialityQA. We expect the generous evaluation with various answer options in TQA dataset to level overall scores upward.

% % between existing models trained with multi-task learning to reveal passage evidence in embeddings, RFiD~\cite{acl/WangY023/RFiD} outperforms EvidentialityQA~\cite{naacl/Asai0H22/EvidentialityQA} by directly adding learnable embeddings. Furthermore, MGFiD improved by 1.4\% over RFiD on NQ and by 0.3\% on TQA. We specifically attribute the performance improvement to the direct delivery of evidence via anchor vector. 

\begin{table}[t]\small
\centering

\begin{tabular}{lcc}
\toprule
Model & R@1 & AUC \\ \midrule
DPR & 49.9 & - \\ \midrule
FiD-KD (cross-attention) &  58.6 & - \\ \midrule
MGFiD (Passage ranker) & \textbf{62.2} & \textbf{0.82} \\
\hspace{6pt}* w/ Cross-entropy $\mathcal{L}_{\text{sentence}}$ & 60.9 & 0.70 \\
\hspace{6pt}* w/o $\mathcal{L}_{\text{sentence}}$ & 61.7 & - \\
\bottomrule
\end{tabular}
\caption{Effectiveness of the proposed method for ranking and classification tasks on the NQ dev dataset. We report MGFiD trained with labels generated by MythoMax. The AUC metric is only reported for MGFiD variants that include the sentence classifier.}
\label{tab:recall_auc}
\end{table}

% Model 대신 Retreiver | Reranker
% 왼쪽 정렬

% \begin{tabular}{c|cc}
% \toprule
% Model & R@1 & AUC \\ \midrule
% DPR & 49.9 & - \\ \midrule
% FiD (cross-attention) &  58.6 & - \\ \midrule
% MGFiD (w/ Pas\mathcal{L}_{\text{sent}}anker) & 61.7& - \\
% \multicolumn{1}{l|}{\hspace{6pt}$+\mathcal{L}_{\text{\mathcal{L}_{\text{sent}} \text{w/ Cross-entropy}$} & 60.9 & 0.70 \\
% \multicolumn{1}{l|}{\hspace{6pt}$+\mathcal{L}_{\text{sent}} \text{w/ Focal loss}$} & \textbf{62.2} & \textbf{0.82} \\
% \multicolumn{1}{l|}{\hspace{6pt}$+\mathcal{L}_{\text{sent.}} \text{w/ Cross-entropy}$} & 60.9 & 0.70 \\
% \multicolumn{1}{l|}{\hspace{6pt}$+\mathcal{L}_{\text{sent.}} \text{w/ Focal loss}$} & \textbf{62.2} & \textbf{0.82} \\

\subsection{In-depth Analysis}\label{sec:exp_in-depth}
\noindent

% fig_filtering.tex
\begin{figure}[t]
    \centering
    \begin{subfigure}[b]{0.2375\textwidth} % Adjust this value as needed to fit within your column width
        \centering
        \includegraphics[width=\linewidth]{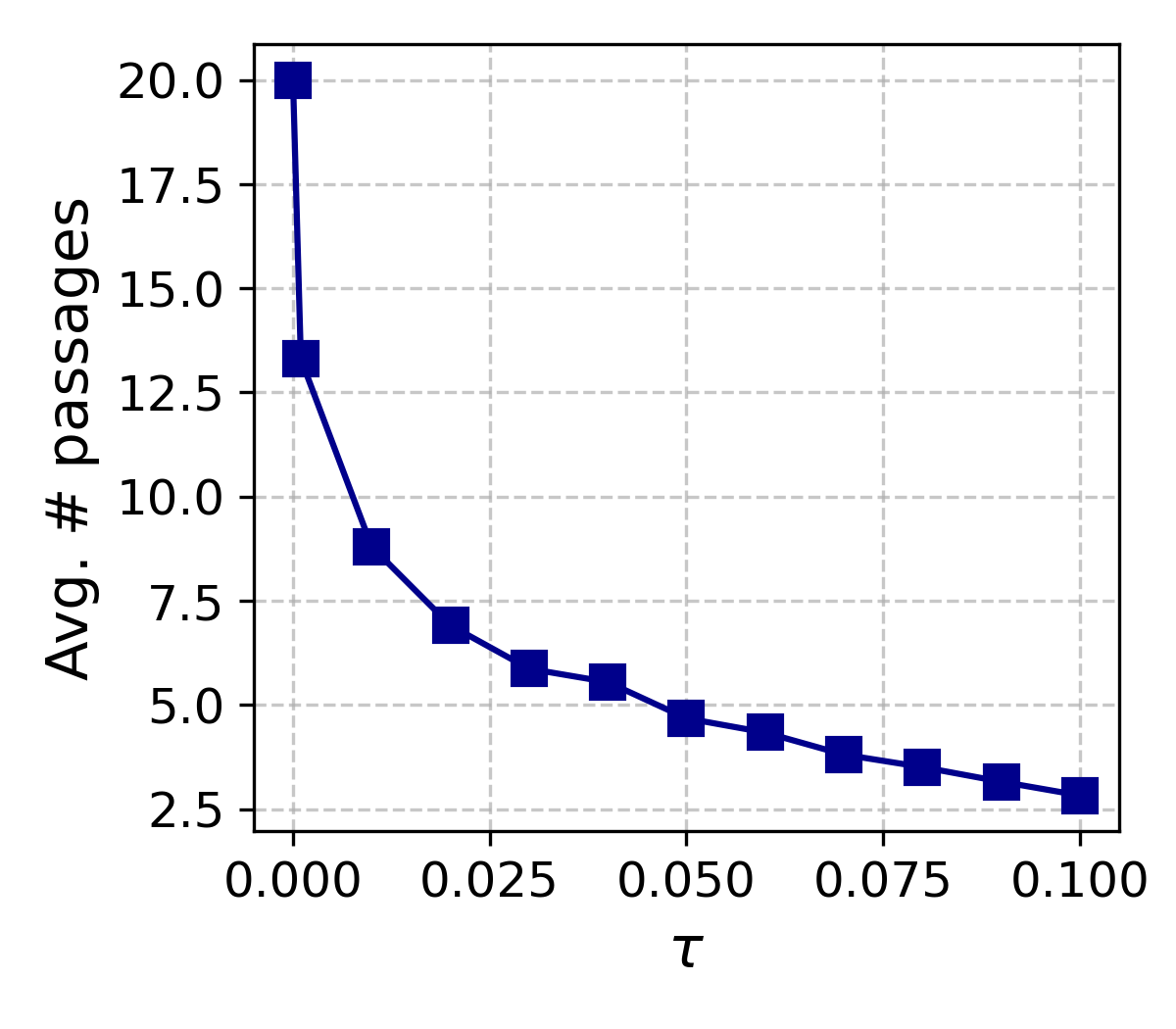}
        \caption{}
        \label{fig:first_graph}
    \end{subfigure}
    \begin{subfigure}[b]{0.2375\textwidth} % Adjust this value as needed to fit within your column width
        \centering
        \includegraphics[width=\linewidth]{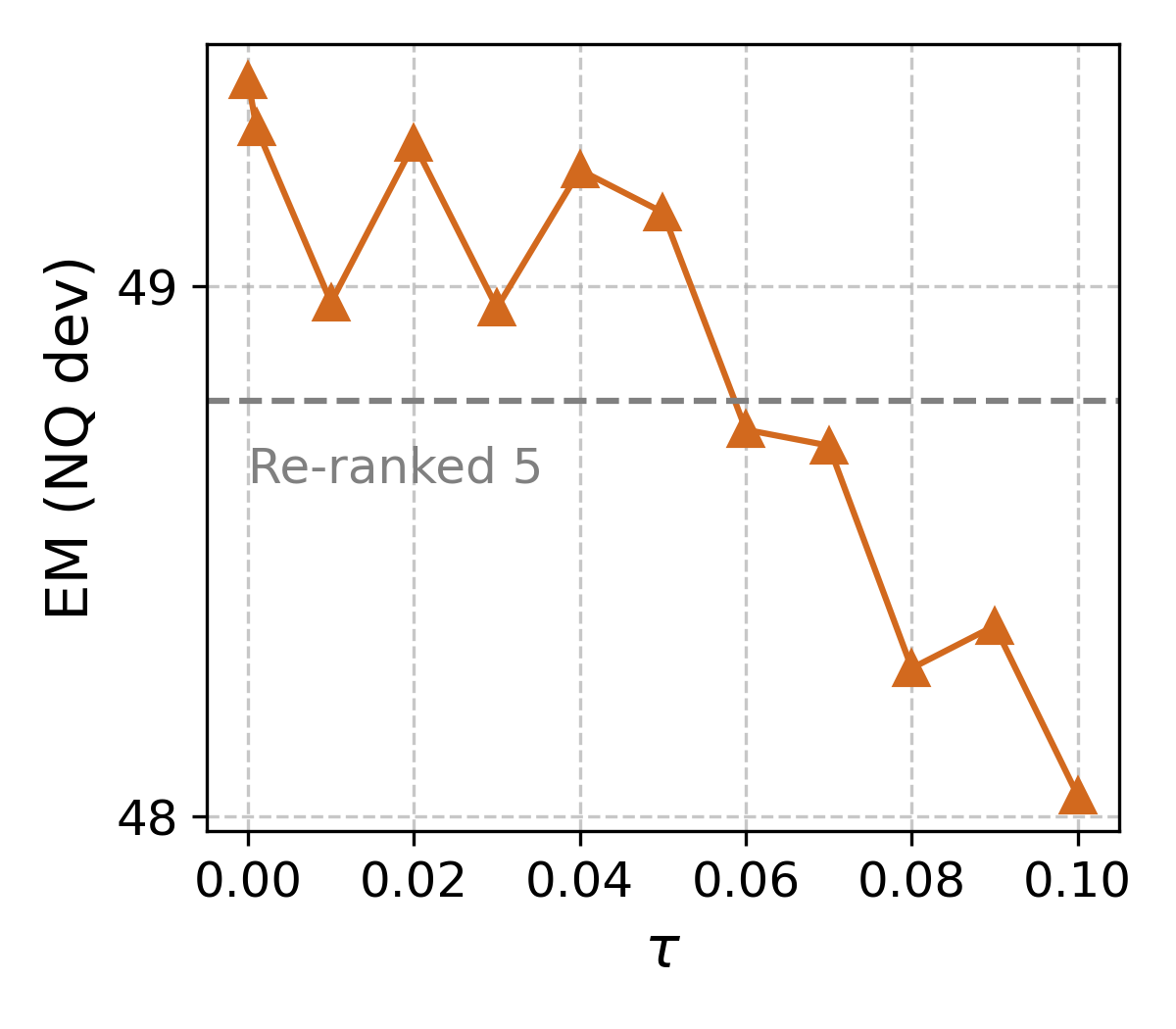} % Replace with the actual file for the second graph
        \caption{}
        \label{fig:second_graph}
    \end{subfigure}
    \caption{(a) The average number of passages provided to the decoder as a function of $\tau$. (b) The effectiveness of varying $\tau$. We utilized the NQ dev dataset and the best checkpoint of MGFiD. When $\tau = 0.05$, MGFiD significantly outperforms using a constant number of 5 re-ranked passages with fewer passages.}
    \label{fig:filtering_percentage}
\vskip -0.1in
\end{figure}

\noindent
\textbf{Ranking \& classification performance.} In Table~\ref{tab:recall_auc}, we measured the outputs of the evidence ranker and sentence classifier as Recall and AUC score, to evaluate our model's ability to identify evidence paragraphs and supporting sentences. (i) The passage ranking score can be implicitly measured by the decoder's cross-attention score. The cross-attention score of each document is calculated by summing the cross-attention scores of the tokens. In this way, the Recall@1 score improved by 17.2\% compared to the DPR retriever. (ii) The improvement is even higher for the passage ranker with explicit ranking capability. When trained with MythoMax labels, it shows an outstanding 5.1\% improvement over the re-ranking result using the cross-attention score in FiD. This suggests that it is more effective to add a module that specializes in determining evidence rather than relying on the cross-attention of the decoder. (iii) When sentence classification is applied, Recall@1 improves even more. Using the focal loss for better classification of imbalanced sentence labels, the AUC score improved to 0.82, and the Recall@1 score reached 62.2. This suggests that emphasizing the embedding of important sentences also helps to distinguish supportive passages.

%, which has to find evidence and generate answers at the same time

\noindent
\textbf{Efficiency via passage pruning.} Figure~\ref{fig:filtering_percentage} shows the number of passages used by the decoder and the effectiveness depending on the pruning threshold $\tau$. It takes all 20 passages when no pruning is applied, \ie, $\tau=0$. Increasing $\tau$ to 0.05 results in a small performance drop even if the number of passages drops drastically below 5. It is worth noting that the performance is much higher than simply using the top-5 passages among the re-ranked passages. Since the concatenation of all encoded tokens causes high computational cost~\cite{sigir/HofstatterC0Z23/FiD-Light}, it helps to avoid a significant performance drop while reducing the decoding overhead.

\begin{table}[t]\small
\centering
\begin{tabular}{cccc|cc}
\toprule
$\mathcal{L}_{\text{ranking}}$ & $\mathcal{L}_{\text{sent}}$ & $e_{\text{anchor}}$ & $\tau$ & NQ & TQA \\ \midrule
\checkmark & \checkmark & \checkmark & 0.05 & \underline{49.1} & 67.7 \\
\checkmark & \checkmark & \checkmark & \textit{top}-$5$ & 48.8 & - \\ \midrule
listwise & \checkmark & \checkmark & $\times$ & \textbf{49.4} & \underline{67.9} \\
listwise & \checkmark & $\times$ & $\times$ & 48.9 & \underline{67.9} \\
$\times$ & \checkmark & $\times$ & $\times$ & 48.1 & \underline{67.9} \\
listwise & $\times$ & $\times$ & $\times$ & 48.8 & 67.8 \\
pointwise & $\times$ & $\times$ & $\times$ & 48.3 & 67.6 \\
$\times$ & $\times$ & $\times$ & $\times$ & 47.8 & 67.5 \\ \bottomrule
\end{tabular}
\caption{Ablation study on the impact of multi-task learning and Threshold-based masking. Note that we are reporting for seed 0 in this result.}
\label{tab:ablation_study}
% \vspace{-3mm}
\vskip -0.1in
\end{table}

\vspace{0.5mm}
\noindent
\textbf{Ablation study.} Table~\ref{tab:ablation_study} shows an ablation study on our different methods. (i) Listwise loss for multi-task learning achieves 0.5\%p higher accuracy than the point-wise loss on the NQ dataset. This implies that listwise is a more reasonable approach due to the structure of FiD, which concatenates multiple passage embeddings and utilizes them at once. (ii) The anchor vector provides core sentence-level information by adding up the anchor vector to the [BOS] token. With this additional information, the accuracy improved by 0.2\%p on the NQ dataset. (iii) When $\mathcal{L}_\text{passage}$ and $\mathcal{L}_\text{sentence}$ are used, the accuracy reaches a peak of 49.4 and 67.9 for the NQ and TQA, respectively. This suggests that multi-granularity can help performance by obtaining more evidentiality. (iv) When $\tau$ is 0.05, the average number of passages used in the decoder is 4.8 in the NQ dataset. Pruned MGFiD gets 0.3\%p higher than using the fixed top-5 re-ranked passages, suggesting that it is more effective to utilize only the supportive passages for each question.

\begin{table}[t]\small
\centering
\begin{tabular}{c|cc|cc}
\toprule
\begin{tabular}[c]{@{}c@{}}Evidence\\ label\end{tabular} & \begin{tabular}[c]{@{}c@{}}NQ dev\\ (EM)\end{tabular}    & \# pos. & \begin{tabular}[c]{@{}c@{}}TQA dev\\ (EM)\end{tabular} & \# pos. \\ \midrule
-                                                        & 47.8 \scriptsize $\pm$ 0.16 & -       & 67.4 \scriptsize $\pm$ 0.12 & -       \\ \midrule
Ans. span                                              & 48.5 \scriptsize $\pm$ 0.21 & 4.5     & 67.7 \scriptsize $\pm$ 0.16     & 8.9     \\
ChatGPT                                                  & \textbf{48.9} \scriptsize $\pm$ 0.13 & 4.0     & 67.7 \scriptsize $\pm$ 0.20      & 8.3 \\
MythoMax                                                 & 48.8 \scriptsize $\pm$ 0.23 & 2.8     & \textbf{67.8} \scriptsize $\pm$ 0.18   & 6.5     \\ \bottomrule
\end{tabular}
\caption{Model performance with different evidence labels. \# pos. denotes the average number of positive labels in top-20 passages.}
\label{tab:label_comparison}
% \vspace{-3mm}
\vskip -0.1in
\end{table}

\subsection{Effectiveness of Evidence Labels}\label{sec:exp_label_performance}

Table~\ref{tab:label_comparison} shows the experiment results of FiD with only passage-level evidence learning, \ie, $\mathcal{L}_\text{passage}$. We use three different labels for passage re-ranking: Ans. span, which checks if the answer span is included, and labels filtered by ChatGPT, and MythoMax. (i) FiD without multi-task learning significantly underperforms on both datasets compared to the others, trained with the additional passage re-ranking. These results suggest that it is insufficient to implicitly let the reader determine evidence without additional ranking information. (ii) The models trained with LLM-generated labels for passage re-ranking outperform those trained with answer span presence as a label, improving by up to 0.4 on the NQ dataset. This suggests that mislabeled spurious passages act as noisy data when the answer spans are used as a determinant of labeling, thereby leading to sub-optimal results. (iii) We note that the performance using the MythoMax label is not significantly different from the performance using the ChatGPT label. This suggests that our framework can effectively determine evidence regardless of the size of the LLMs.
% In other words, the LLMs available for filtering the evidence labels are not very limited.

\subsection{Effectiveness by the Number of Passages}\label{sec:exp_varying_num_passages}

\begin{figure}[t]
\includegraphics[width=1.0\linewidth]{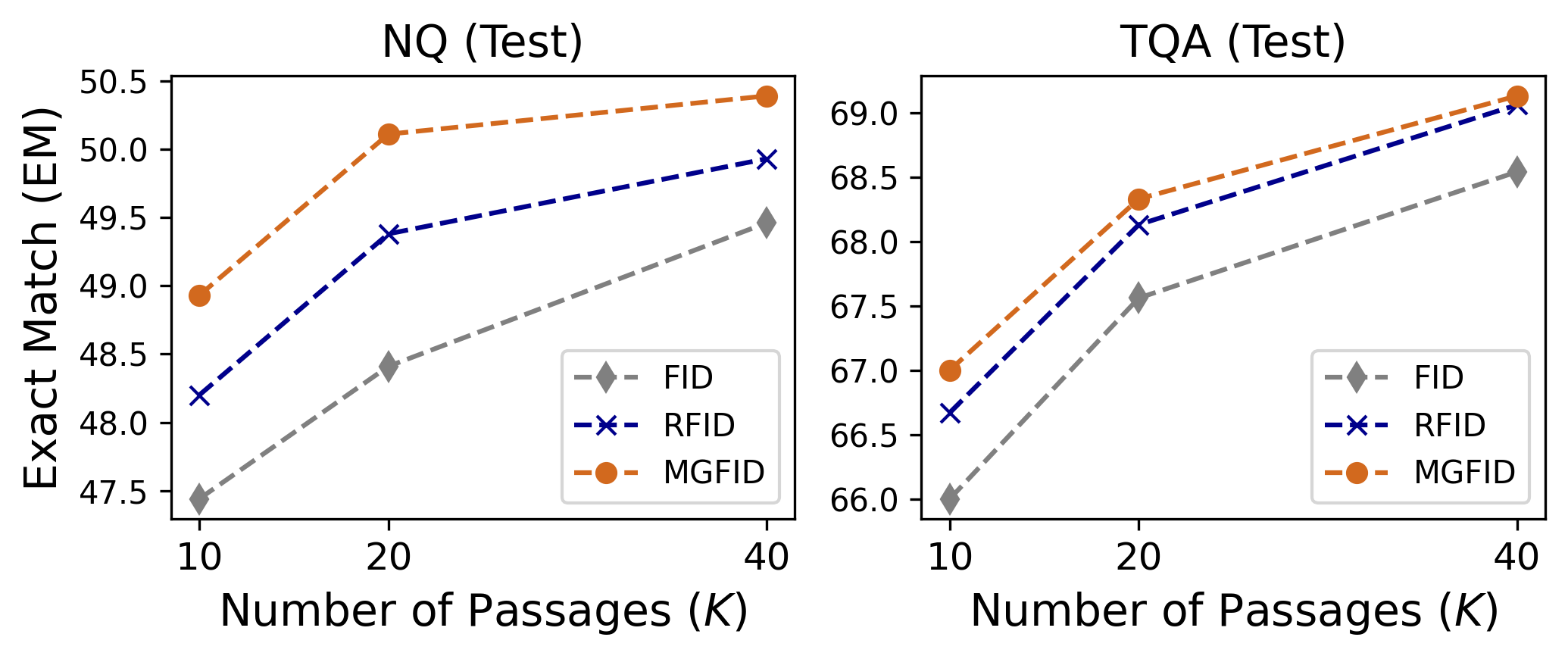}
\caption{Effectiveness of FiD-KD~\cite{iclr/IzacardG21/FiD-KD}, RFiD~\cite{acl/WangY023/RFiD}, and MGFiD varying the number of passages used in the encoder, \ie, $K$.}\label{fig:fig_varying_passages}
\vskip -0.1in
\end{figure}

Figure~\ref{fig:fig_varying_passages} illustrates the performance of MGFiD and two baseline models on the NQ and TQA test sets with varying numbers of passages used by the encoder, \ie, $K$. We trained each model using the top-$K$ passages. Our findings reveal that performance is enhanced with more passages for all models, aligning with the aggregating capability of the FiD architecture noted by ~\citet{eacl/IzacardG21/FiD}. Second, MGFiD consistently outperforms the baselines across different numbers of passages (10, 20, and 40), highlighting the significance of the capability to discern supportive passages. Lastly, the efficacy of evidence-based multi-task learning, as utilized by MGFiD and RFiD~\cite{acl/WangY023/RFiD}, is most significant with fewer documents, \ie, $K=10$. This observation is counterintuitive to the expectation that filtering spurious passages becomes more critical as the number of passages increases. We interpret this to suggest that increasing the number of passages may have a similar effect as increasing the batch size~\cite{naacl/QuDLLRZDWW21/RocketQA}, whereas the multi-task learning can efficiently achieve high performance even with smaller batch sizes. We leave a more detailed analysis to future work.

\section{Conclusion}\label{sec:conclusion}

This paper presents the Multi-Granularity Guided Fusion-in-Decoder (MGFiD), a novel reader for managing evidence across multiple granularities. Addressing the prevalent challenges of misleading passages and sentences, MGFiD synergies coarse-level passage re-ranking with fine-level sentence classification. We also incorporate LLMs to enhance the quality of heuristic labels. Moreover, MGFiD capitalizes on its multi-granularity evidence by constructing an anchor vector that guides the decoder toward significant evidence and employs passage pruning to enhance decoding efficiency. Our empirical results demonstrate that MGFiD using multi-granularity contexts achieves significant advancements over baseline models.

%by providing better insight into combining evidence across multiple passages.

% AI대학원, 나비효과, 오픈도메인, AGC
\section*{Acknowledgments} 
This work was supported by Institute of Information \& communications Technology Planning \& Evaluation (IITP) grant funded by the Korea government (MSIT) (No. 2019-0-00421, 2022-0-00680-003, 2022-0-01045, and RS-2023-00219919).
\section*{Limitations}
We briefly describe the limitations of our method. (i) LLM filtering methods are limited to extractive QA for the current setting. (ii) There needs to be validation on more passages. (iii) Marginal improvement on TQA dataset.

\noindent
\textbf{Limitation of LLM labels.} Because our label filtering method is based on answer span, it is still quite limited to the extractive task. However, the criterion for silver labels is not necessarily answer span, and we have shown in the paper that the filtering task does not necessarily require expensive models. This means that for relatively low $K$, it is available to perform on all the retrieved results. The fact that harsh filtering by MythoMax worked even with fewer labels means that the multi-task does not necessarily require many labels.

\noindent
\textbf{A large number of passages}. We do not report results using a large number of passages, \eg, 100, and a bigger backbone model, \ie, T5-large, due to the computational cost. Previous research has shown that using more passages increases the probability that the passage set contains evidence and thus improves performance. We also found in our experiments that the standard deviation of the NQ dataset is large, depending on the seed. This was true for all of the baseline models we reproduced. Although we compared our model and the baseline with five seeds, it would be desirable to validate additional seeds to further examine generalizability.

\noindent
\textbf{Marginal improvement on TQA.} The performance improvement on TQA is marginal compared to that of NQ. We can assume that the multi-task learning to identify supportive context is less effective for TQA because it has numerous answer candidates and passages regarding evidence are present. It is thus relatively easy to get an EM score. However, we still need to analyze this further.

% As our label filtering method is based on answer span, it is still quite limited to extractive task. 그러나, silver label의 기준은 꼭 answer span이 아니어도 되며, 우리는 논문을 통해 filtering task에 꼭 비싼 모델이 필요하지 않음을 보였다. 즉, relatively 낮은 k에 대해서는 모든 검색 결과에 대해 수행하는 것도 가능하며, MythoMax의 harsh한 filtering이 label 수를 적게 만들었어도 효과가 있었음은 해당 multi-task가 꼭 많은 label이 필요하지는 않다는 것을 의미한다.

% 다양한 K와 backbone 모델에 대한 검증이 필요하다. 하지만 기존 연구에서 보이듯이, 검색 문서를 많이 사용할 수록 문서 집합에 근거가 포함될 확률이 올라가고, 이에 따라 성능이 개선될 것으로 기대할 수 있다. 또한 많은 문서를 사용할 수록 문서를 분별하여 읽는 것의 중요도가 올라가므로, 우리 방법론의 효과가 더욱 커질 수 있다. computing 비용으로 인해 reporting 못 했지만 향후 이와 관련된 검증을 수행할 예정이다. 그 보다 중요한 것은, 우리는 실험을 통해 NQ dataset의 seed에 따른 표준편차가 작지 않음을 발견하였다. 이것은 재현을 수행한 모든 baseline 모델에 해당하는 사항이었다. 5번의 시드로 baseline과 비교하였지만, 더욱 많은 시드를 활용하여 검증을 진행하는 것이 바람직하다.
\section*{Ethics Statement}

The ethical guidelines of ACL are fully respected in this work. We have utilized scientific resources available for research under liberal licenses. Our use of these tools is consistent with their intended applications.

% Entries for the entire Anthology, followed by custom entries
\bibliographystyle{acl_natbib}
\bibliography{references}

\begin{thebibliography}{26}
\expandafter\ifx\csname natexlab\endcsname\relax\def\natexlab#1{#1}\fi

\bibitem[{Asai et~al.(2022)Asai, Gardner, and
  Hajishirzi}]{naacl/Asai0H22/EvidentialityQA}
Akari Asai, Matt Gardner, and Hannaneh Hajishirzi. 2022.
\newblock \href {https://doi.org/10.18653/v1/2022.naacl-main.162}
  {Evidentiality-guided generation for knowledge-intensive {NLP} tasks}.
\newblock In \emph{NAACL}, pages 2226--2243.

\bibitem[{Bradley(1997)}]{pr/Bradley97/AUC}
Andrew~P. Bradley. 1997.
\newblock \href {https://doi.org/10.1016/S0031-3203(96)00142-2} {The use of the
  area under the {ROC} curve in the evaluation of machine learning algorithms}.
\newblock \emph{Pattern Recognit.}, 30:1145--1159.

\bibitem[{Chen et~al.(2017)Chen, Fisch, Weston, and
  Bordes}]{acl/ChenFWB17/DrQA}
Danqi Chen, Adam Fisch, Jason Weston, and Antoine Bordes. 2017.
\newblock \href {https://doi.org/10.18653/v1/P17-1171} {Reading wikipedia to
  answer open-domain questions}.
\newblock In \emph{ACL}, pages 1870--1879.

\bibitem[{de~Jong et~al.(2023)de~Jong, Zemlyanskiy, Ainslie, FitzGerald,
  Sanghai, Sha, and Cohen}]{acl/JongZAFSSC23/FiDO}
Michiel de~Jong, Yury Zemlyanskiy, Joshua Ainslie, Nicholas FitzGerald, Sumit
  Sanghai, Fei Sha, and William~W. Cohen. 2023.
\newblock \href {https://doi.org/10.18653/v1/2023.findings-acl.732} {Fido:
  Fusion-in-decoder optimized for stronger performance and faster inference}.
\newblock In \emph{Findings of the ACL}, pages 11534--11547.

\bibitem[{Guu et~al.(2020)Guu, Lee, Tung, Pasupat, and
  Chang}]{icml/GuuLTPC20/REALM}
Kelvin Guu, Kenton Lee, Zora Tung, Panupong Pasupat, and Ming{-}Wei Chang.
  2020.
\newblock \href {http://proceedings.mlr.press/v119/guu20a.html} {Retrieval
  augmented language model pre-training}.
\newblock In \emph{ICML}, pages 3929--3938.

\bibitem[{Hofst{\"{a}}tter et~al.(2023)Hofst{\"{a}}tter, Chen, Raman, and
  Zamani}]{sigir/HofstatterC0Z23/FiD-Light}
Sebastian Hofst{\"{a}}tter, Jiecao Chen, Karthik Raman, and Hamed Zamani. 2023.
\newblock \href {https://doi.org/10.1145/3539618.3591687} {Fid-light: Efficient
  and effective retrieval-augmented text generation}.
\newblock In \emph{SIGIR}, pages 1437--1447.

\bibitem[{Izacard and Grave(2021{\natexlab{a}})}]{iclr/IzacardG21/FiD-KD}
Gautier Izacard and Edouard Grave. 2021{\natexlab{a}}.
\newblock \href {https://openreview.net/forum?id=NTEz-6wysdb} {Distilling
  knowledge from reader to retriever for question answering}.
\newblock In \emph{ICLR}.

\bibitem[{Izacard and Grave(2021{\natexlab{b}})}]{eacl/IzacardG21/FiD}
Gautier Izacard and Edouard Grave. 2021{\natexlab{b}}.
\newblock \href {https://doi.org/10.18653/v1/2021.eacl-main.74} {Leveraging
  passage retrieval with generative models for open domain question answering}.
\newblock In \emph{EACL}, pages 874--880.

\bibitem[{Joshi et~al.(2017)Joshi, Choi, Weld, and
  Zettlemoyer}]{acl/JoshiCWZ17/TQA}
Mandar Joshi, Eunsol Choi, Daniel~S. Weld, and Luke Zettlemoyer. 2017.
\newblock \href {https://doi.org/10.18653/v1/P17-1147} {Triviaqa: {A} large
  scale distantly supervised challenge dataset for reading comprehension}.
\newblock In \emph{ACL}, pages 1601--1611.

\bibitem[{Ju et~al.(2022)Ju, Yu, Zhao, Zhang, and Ye}]{emnlp/Ju00Z022/GRAPE}
Mingxuan Ju, Wenhao Yu, Tong Zhao, Chuxu Zhang, and Yanfang Ye. 2022.
\newblock \href {https://doi.org/10.18653/v1/2022.findings-emnlp.13} {Grape:
  Knowledge graph enhanced passage reader for open-domain question answering}.
\newblock In \emph{Findings of EMNLP}, pages 169--181.

\bibitem[{Karpukhin et~al.(2020)Karpukhin, Oguz, Min, Lewis, Wu, Edunov, Chen,
  and Yih}]{emnlp/KarpukhinOMLWEC20/DPR}
Vladimir Karpukhin, Barlas Oguz, Sewon Min, Patrick S.~H. Lewis, Ledell Wu,
  Sergey Edunov, Danqi Chen, and Wen{-}tau Yih. 2020.
\newblock \href {https://doi.org/10.18653/v1/2020.emnlp-main.550} {Dense
  passage retrieval for open-domain question answering}.
\newblock In \emph{EMNLP}, pages 6769--6781.

\bibitem[{Kingma and Ba(2015)}]{iclr/KingmaB14/adam}
Diederik~P. Kingma and Jimmy Ba. 2015.
\newblock \href {http://arxiv.org/abs/1412.6980} {Adam: {A} method for
  stochastic optimization}.
\newblock In \emph{ICLR}.

\bibitem[{Kwiatkowski et~al.(2019)Kwiatkowski, Palomaki, Redfield, Collins,
  Parikh, Alberti, Epstein, Polosukhin, Devlin, Lee, Toutanova, Jones, Kelcey,
  Chang, Dai, Uszkoreit, Le, and Petrov}]{tacl/KwiatkowskiPRCP19/NQ}
Tom Kwiatkowski, Jennimaria Palomaki, Olivia Redfield, Michael Collins,
  Ankur~P. Parikh, Chris Alberti, Danielle Epstein, Illia Polosukhin, Jacob
  Devlin, Kenton Lee, Kristina Toutanova, Llion Jones, Matthew Kelcey,
  Ming{-}Wei Chang, Andrew~M. Dai, Jakob Uszkoreit, Quoc Le, and Slav Petrov.
  2019.
\newblock \href {https://doi.org/10.1162/tacl\_a\_00276} {Natural questions: a
  benchmark for question answering research}.
\newblock \emph{Trans. Assoc. Comput. Linguistics}, pages 452--466.

\bibitem[{Lakhotia et~al.(2021)Lakhotia, Paranjape, Ghoshal, Yih, Mehdad, and
  Iyer}]{emnlp/LakhotiaPGYMI21/FiD-Ex}
Kushal Lakhotia, Bhargavi Paranjape, Asish Ghoshal, Scott Yih, Yashar Mehdad,
  and Srini Iyer. 2021.
\newblock \href {https://doi.org/10.18653/v1/2021.emnlp-main.301} {Fid-ex:
  Improving sequence-to-sequence models for extractive rationale generation}.
\newblock In \emph{EMNLP}, pages 3712--3727.

\bibitem[{Lee et~al.(2022)Lee, Kedia, Lee, Paranjape, Manning, and
  Woo}]{emnlp/LeeKLPMW22/YONO}
Haejun Lee, Akhil Kedia, Jongwon Lee, Ashwin Paranjape, Christopher~D. Manning,
  and Kyoung{-}Gu Woo. 2022.
\newblock \href {https://doi.org/10.18653/v1/2022.emnlp-main.198} {You only
  need one model for open-domain question answering}.
\newblock In \emph{EMNLP}, pages 3047--3060.

\bibitem[{Lee et~al.(2019)Lee, Chang, and Toutanova}]{acl/LeeCT19/ORQA}
Kenton Lee, Ming{-}Wei Chang, and Kristina Toutanova. 2019.
\newblock \href {https://doi.org/10.18653/v1/p19-1612} {Latent retrieval for
  weakly supervised open domain question answering}.
\newblock In \emph{ACL}, pages 6086--6096.

\bibitem[{Lewis et~al.(2020)Lewis, Perez, Piktus, Petroni, Karpukhin, Goyal,
  K{\"{u}}ttler, Lewis, Yih, Rockt{\"{a}}schel, Riedel, and
  Kiela}]{nips/LewisPPPKGKLYR020/RAG}
Patrick S.~H. Lewis, Ethan Perez, Aleksandra Piktus, Fabio Petroni, Vladimir
  Karpukhin, Naman Goyal, Heinrich K{\"{u}}ttler, Mike Lewis, Wen{-}tau Yih,
  Tim Rockt{\"{a}}schel, Sebastian Riedel, and Douwe Kiela. 2020.
\newblock \href
  {https://proceedings.neurips.cc/paper/2020/hash/6b493230205f780e1bc26945df7481e5-Abstract.html}
  {Retrieval-augmented generation for knowledge-intensive {NLP} tasks}.
\newblock In \emph{NeurIPS}.

\bibitem[{Lin et~al.(2017)Lin, Goyal, Girshick, He, and
  Doll{\'{a}}r}]{iccv/LinGGHD17/focal}
Tsung{-}Yi Lin, Priya Goyal, Ross~B. Girshick, Kaiming He, and Piotr
  Doll{\'{a}}r. 2017.
\newblock \href {https://doi.org/10.1109/ICCV.2017.324} {Focal loss for dense
  object detection}.
\newblock In \emph{ICCV}, pages 2999--3007.

\bibitem[{Liu et~al.(2023)Liu, He, Guo, Chen, Hui, and
  Zhou}]{cikm/0001HGCHZ23/GranCATs}
Meizhen Liu, Jiakai He, Xu~Guo, Jianye Chen, Siu~Cheung Hui, and Fengyu Zhou.
  2023.
\newblock \href {https://doi.org/10.1145/3583780.3614896} {Grancats:
  Cross-lingual enhancement through granularity-specific contrastive adapters}.
\newblock In \emph{CIKM}, pages 1461--1471.

\bibitem[{Nogueira and Cho(2019)}]{corr/abs-1901-04085/monoBERT}
Rodrigo Nogueira and Kyunghyun Cho. 2019.
\newblock \href {http://arxiv.org/abs/1901.04085} {Passage re-ranking with
  bert}.
\newblock \emph{CoRR}.

\bibitem[{Qu et~al.(2021)Qu, Ding, Liu, Liu, Ren, Zhao, Dong, Wu, and
  Wang}]{naacl/QuDLLRZDWW21/RocketQA}
Yingqi Qu, Yuchen Ding, Jing Liu, Kai Liu, Ruiyang Ren, Wayne~Xin Zhao, Daxiang
  Dong, Hua Wu, and Haifeng Wang. 2021.
\newblock \href {https://doi.org/10.18653/v1/2021.naacl-main.466} {Rocketqa: An
  optimized training approach to dense passage retrieval for open-domain
  question answering}.
\newblock In \emph{NAACL-HLT}, pages 5835--5847.

\bibitem[{Raffel et~al.(2020)Raffel, Shazeer, Roberts, Lee, Narang, Matena,
  Zhou, Li, and Liu}]{JMLR/Raffel2020/T5}
Colin Raffel, Noam Shazeer, Adam Roberts, Katherine Lee, Sharan Narang, Michael
  Matena, Yanqi Zhou, Wei Li, and Peter~J. Liu. 2020.
\newblock \href {http://jmlr.org/papers/v21/20-074.html} {Exploring the limits
  of transfer learning with a unified text-to-text transformer}.
\newblock \emph{J. Mach. Learn. Res.}, 21:140:1--140:67.

\bibitem[{Sun et~al.(2023)Sun, Yan, Ma, Wang, Ren, Chen, Yin, and
  Ren}]{DBLP:conf/emnlp/RankGPT}
Weiwei Sun, Lingyong Yan, Xinyu Ma, Shuaiqiang Wang, Pengjie Ren, Zhumin Chen,
  Dawei Yin, and Zhaochun Ren. 2023.
\newblock \href {https://aclanthology.org/2023.emnlp-main.923} {Is chatgpt good
  at search? investigating large language models as re-ranking agents}.
\newblock In \emph{EMNLP}, pages 14918--14937.

\bibitem[{Wagner(2010)}]{lre/Wagner10/NLTK}
Wiebke Wagner. 2010.
\newblock \href {https://doi.org/10.1007/s10579-010-9124-x} {Steven bird, ewan
  klein and edward loper: Natural language processing with python, analyzing
  text with the natural language toolkit - o'reilly media, beijing, 2009,
  {ISBN} 978-0-596-51649-9}.
\newblock \emph{Lang. Resour. Evaluation}, 44:421--424.

\bibitem[{Wang et~al.(2023)Wang, Yu, and Zhang}]{acl/WangY023/RFiD}
Cunxiang Wang, Haofei Yu, and Yue Zhang. 2023.
\newblock \href {https://doi.org/10.18653/v1/2023.findings-acl.155} {Rfid:
  Towards rational fusion-in-decoder for open-domain question answering}.
\newblock In \emph{Findings of ACL}, pages 2473--2481.

\bibitem[{Yu et~al.(2022)Yu, Zhu, Fang, Yu, Wang, Xu, Ren, Yang, and
  Zeng}]{acl/Yu0F0WXRY022/KG-FiD}
Donghan Yu, Chenguang Zhu, Yuwei Fang, Wenhao Yu, Shuohang Wang, Yichong Xu,
  Xiang Ren, Yiming Yang, and Michael Zeng. 2022.
\newblock \href {https://doi.org/10.18653/v1/2022.acl-long.340} {Kg-fid:
  Infusing knowledge graph in fusion-in-decoder for open-domain question
  answering}.
\newblock In \emph{ACL}, pages 4961--4974.

\end{thebibliography}

\appendix

\section{Appendix}

% \subsection{Evidence Labeling}\label{sec:app_el}

% To filter out spurious passages, we utilize well-known LLMs, ChatGPT~\footnote{https://chat.openai.com/chat}, and MythoMax-13B~\footnote{https://huggingface.co/TheBloke/MythoMax-L2-13B-GPTQ}, which is relatively small and can be easily used in the local GPU. For ChatGPT, we fix the temperature to 0, and for MythoMax, we do not use sampling. We use all the answer candidates in NQ; however, in the TQA dataset, which has many more answer candidates than NQ, we only collected answers in the top 20 passages for efficient prompting.
% Figure~\ref{fig:prompt} shows an LLM prompt for the labeling task. For each passage, we instructed LLM to give a short answer of 'yes' or 'no' to the passage's relevance to the question, given the \{question, passage(title and context), answers\} triplets. These answers are converted to 1s and 0s and then appended to the training dataset as evidence labels. 

\begin{figure}[h]
\includegraphics[width=1.0\linewidth]{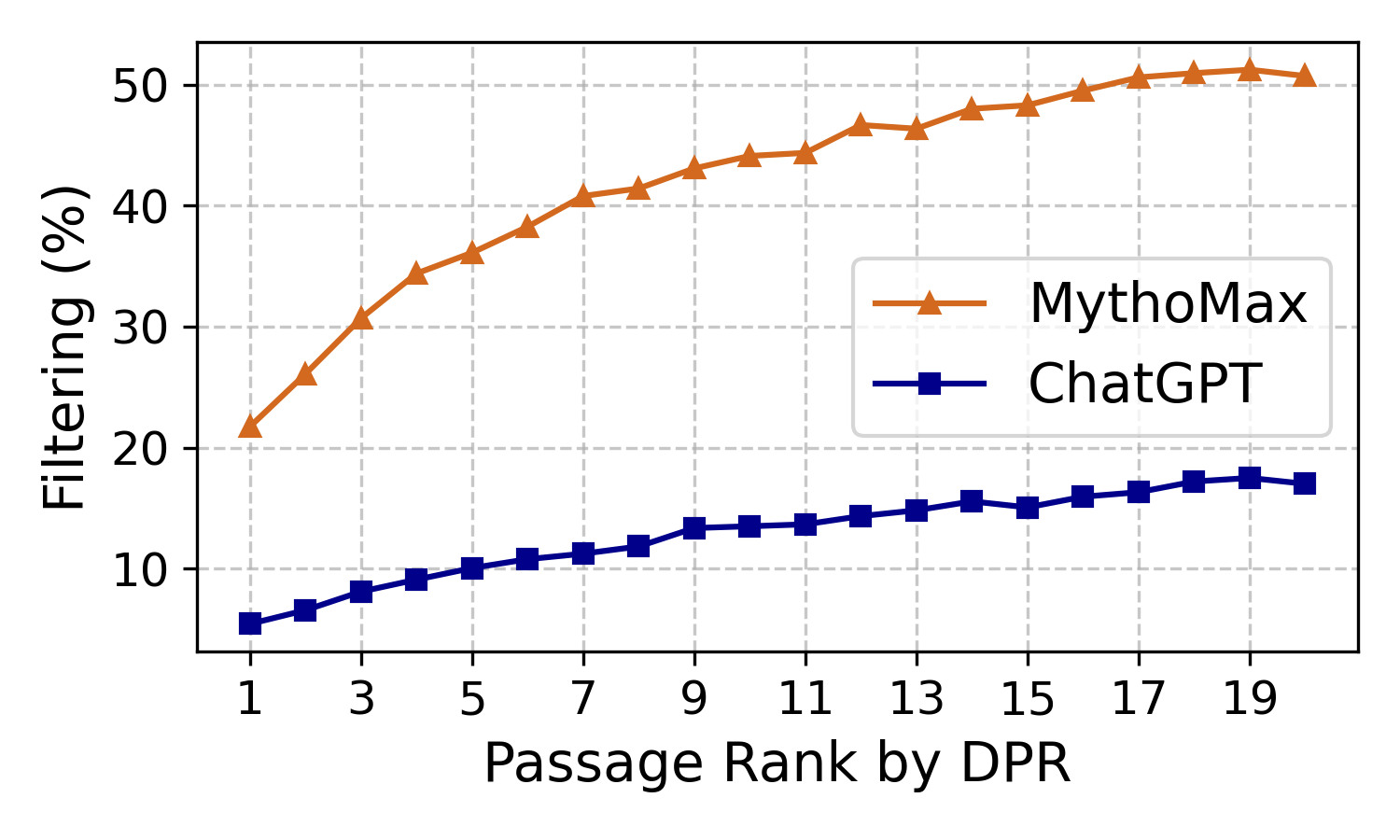}
\caption{Filtering percentage by rank. Both MythoMax and ChatGPT show more than 10\% and 40\% filtering ratios at top-10 ranking results. This suggests that both systems are doing the task reasonably, as the rankings in DPR are likely related to how well the content semantically matches.}\label{fig:filtering}
\vskip -0.1in
\end{figure}

\subsection{Evaluation on label filtering}\label{sec:exp_eval_on_filtering}

Assuming that the rank provided by the retriever~\cite{iclr/IzacardG21/FiD-KD} represents the contextual relevance of a query to a paragraph, it is reasonable to expect the distribution of desirable supporting passages in the top 20 documents to be asymmetric, with dense at the high ranks and sparse at the low ranks. Figure~\ref{fig:filtering} shows the percentage of passages filtered out (labeled as irrelevant) when passages corresponding to each DPR rank are given to Mythomax and ChatGPT along with a question.
As we expected, both models are more likely to label rank20 passages as irrelevant than rank1 passages. ChatGPT labels very few passages as irrelevant at rank 1, but this increases to almost 20\% as the rank decreases. Mythomax labels over 50\% of passages as irrelevant at low rank. This empirically verifies that LLM's label filtering tendency is consistent with the contextual relevance across ranks.

\begin{figure}[t]
\includegraphics[width=1.0\linewidth]{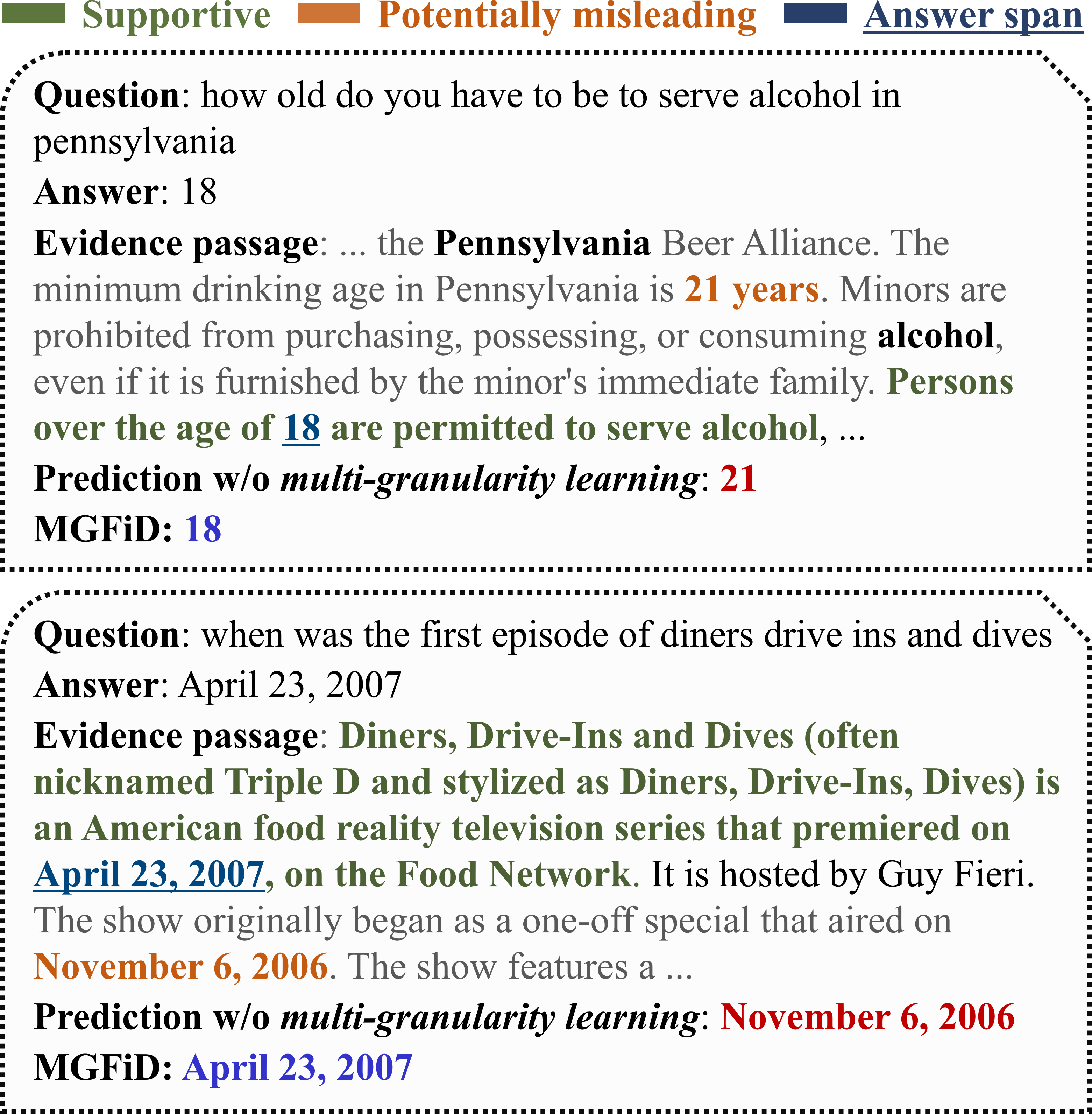}
\caption{More examples that can harm QA systems similar to Figure~\ref{fig:NotEvident}. Two examples show the need to identify which sentence is supportive and which is not. Black \textbf{bold} terms in the passages are overlapped with the question.}
\label{fig:a_notevident}
\vskip -0.1in
\end{figure}

\subsection{Importance of Sentence-level Evidence}\label{sec:app_2}
Existing works only identify which passages are supporting and focus on aggregating evidence across multiple passages. Still, there is a lot of information in the passages that can mislead the model. Figure~\ref{fig:a_notevident} shows that a model trained only on identifying supporting passages, \ie, w/o multi-granularity learning, generated incorrect answers. On the other hand, MGFiD, which learned sentence-level evidence, avoided plausible incorrect answers and generated correct answers.

\end{document}